\documentclass[sigconf]{acmart}

\usepackage{booktabs} 

\usepackage{caption}
\usepackage{epsfig,subfigure}
\usepackage{multirow}
\usepackage{enumitem}
\usepackage{hyperref}
\setitemize{noitemsep,topsep=0pt,parsep=0pt,partopsep=0pt}

\copyrightyear{2018} 
\acmYear{2018} 
\setcopyright{acmcopyright}
\acmConference[MM '18]{2018 ACM Multimedia Conference}{October 22--26, 2018}{Seoul, Republic of Korea}
\acmBooktitle{2018 ACM Multimedia Conference (MM '18), October 22--26, 2018, Seoul, Republic of Korea}
\acmPrice{15.00}
\acmDOI{10.1145/3240508.3240704}
\acmISBN{978-1-4503-5665-7/18/10}

\begin{document}
\title{GestureGAN for Hand Gesture-to-Gesture Translation\\ in the Wild}

\author{Hao Tang$^1$, \quad Wei Wang$^{1,2}$, \quad Dan Xu$^{1,3}$, \quad Yan Yan$^4$, \quad Nicu Sebe$^1$}
\affiliation{%
	\institution{$^1$Department of Information Engineering and Computer Science, University of Trento, Trento, Italy \\
		$^2$Computer Vision Laboratory, \'Ecole Polytechnique F\'ed\'erale de Lausanne, Lausanne, Switzerland \\
		$^3$Department of Engineering Science, University of Oxford, Oxford, United Kingdom \\
		$^4$Department of Computer Science, Texas State University, San Marcos, USA}
{ \{hao.tang,\,niculae.sebe\}@unitn.it,\,wei.wang@epfl.ch,\,danxu@robots.ox.ac.uk,\,y\_y34@txstate.edu}
}

\begin{abstract}
	
Hand gesture-to-gesture translation in the wild is a challenging task since hand gestures can have arbitrary poses, sizes, locations and self-occlusions. 
Therefore, this task requires a high-level understanding of the mapping between the input source gesture and the output target gesture.
To tackle this problem, we propose a novel hand Gesture Generative Adversarial Network (GestureGAN). 
GestureGAN consists of a single generator $G$ and a discriminator $D$, which takes as input a conditional hand image and a target hand skeleton image. 
GestureGAN utilizes the hand skeleton information explicitly, and learns the gesture-to-gesture mapping through two novel losses, the color loss and the cycle-consistency loss. 
The proposed color loss handles the issue of ``channel pollution'' while back-propagating the gradients.
In addition, we present the Fr\'echet ResNet Distance (FRD) to evaluate the quality of generated images.
Extensive experiments on two widely used benchmark datasets demonstrate that the proposed GestureGAN achieves state-of-the-art performance on the unconstrained hand gesture-to-gesture translation task. 
Meanwhile, the generated images are in high-quality and are photo-realistic, allowing them to be used as data augmentation to improve the performance of a hand gesture classifier. 
Our model and code are available at \url{https://github.com/Ha0Tang/GestureGAN}.
	
\end{abstract}



\maketitle

\section{Introduction}

Hand gesture-to-gesture translation in the wild is a task that converts the hand gesture of a given image to a target gesture with a different pose, size and location while preserving the identity information.
This task has many applications, such as human-computer interactions, entertainment, virtual reality and data augmentation.
However, this task is difficult since it needs (i) handling complex backgrounds with different illumination conditions, objects and occlusions; (ii) a high-level semantic understanding of the mapping between the input and output gestures.

Recently, Generative Adversarial Networks (GANs) \cite{goodfellow2014generative} have shown the potential to solve this challenging task. GAN is a generative model based on game theory, which has achieved impressive performance in many applications, such as high-quality image generation \cite{karras2017progressive}, video generation \cite{yan2017skeleton} and audio generation \cite{oord2016wavenet}.
To generate specific kinds of images, videos and audios,  Mirza et al.~\cite{mirza2014conditional} proposed the Conditional GAN (CGAN), which comprises a vanilla GAN and other external information, such as class labels~\cite{choi2017stargan}, text descriptions~\cite{reed2016learning}, images~ \cite{isola2017image} and object keypoints~\cite{reed2016learning}.

In this paper, we focus on the image-to-image translation task using CGAN.
Image-to-image translation tasks can be divided into two types: paired \cite{isola2017image,ma2017pose,siarohin2017deformable} and unpaired \cite{zhu2017unpaired,yi2017dualgan,anoosheh2017combogan,choi2017stargan}.
However, existing image-to-image translation frameworks are inefficient in the multi-domain image-to-image translation task. 
For instance, given $m$ image domains, pix2pix \cite{isola2017image} and  BiCycleGAN \cite{zhu2017toward} need to train $A_m^2{=}m(m{-}1){=}\Theta(m^2)$ models.
CycleGAN~\cite{zhu2017unpaired}, DiscoGAN~\cite{kim2017learning} and DualGAN~\cite{yi2017dualgan} need to train $C_m^2{=}\frac{m(m{-}1)}{2}{=}\Theta(m^2)$ models, or $m(m{-}1)$ generator/discriminator pairs since one model has 2 different generator/discriminator pairs for these methods.
ComboGAN~\cite{anoosheh2017combogan} requires $m{=}\Theta(m)$ models.
StarGAN \cite{choi2017stargan} needs one model. 
However, for some specific image-to-image translation applications  such as hand gesture-to-gesture translation, $m$ could be arbitrary large since gestures in the wild can have arbitrary poses, sizes, appearances, locations and self-occlusions.

To address these limitations, several works have been proposed to generate images based on object keypoints.
For instance, 
Reed et al. \cite{reed2016generating} present an extension of Pixel Convolutional Neural Networks (PixelCNN) to generate images based on keypoints and text description.
Siarohin et al. \cite{siarohin2017deformable} introduce a deformable Generative Adversarial Network for pose-based human image generation.
Ma et al. \cite{ma2017disentangled} propose a two-stage reconstruction pipeline that learns generates novel person images.
However, the aforementioned methods always have the ``channel pollution'' problem that is frequently occurring in generative models such as PG$^2$ \cite{ma2017pose} leading to blurred generated images.
To solve this issue, in this paper, we propose a novel Generative Adversarial Network, \textit{i.e.}, GestureGAN which treats each channel independently.
It allows generating high-quality hand gesture images with arbitrary poses, sizes and locations in the wild, and thus reducing the dependence on environment and pre-processing operations. 
GestureGAN only consists of one generator and one discriminator, taking  a conditional hand gesture image and a target hand skeleton image as inputs.
In addition, to better learn the mapping between inputs and outputs, we propose two novel losses, \textit{i.e.}, color loss and cycle-consistency loss.
Note that the color loss can handle the problem of ``channel pollution'', 
making the generated images sharper and having higher quality.
Furthermore, we propose the Fr\'echet ResNet Distance (FRD), which is a novel evaluation metric to evaluate the generated image of GANs.
Extensive experiments on two public benchmark datasets demonstrate that GestureGAN can generate high-quality images with convincing details. 
Thus, these generated images can augment the training data and improve the performance of  hand gesture classifiers.

Overall, the contributions of this paper are as follows:
\begin{itemize}[leftmargin=*]
	\item We propose a novel Generative Adversarial Network, \textit{i.e.}, GestureGAN,  which can generate target hand gesture with arbitrary poses, sizes and locations in the wild. In addition, we present a novel color loss to learn the hand gesture-to-gesture mappings, handling the problem of ``channel pollution''.
	\item We propose an efficient Fr\'echet ResNet Distance (FRD) metric to evaluate the similarity of the real and generated images, which is more consistent with human judgment.
	FRD measures the similarity between the real image and the generated image in a high-level semantic feature space.
	\item  Qualitative and quantitative results demonstrate the superiority of the proposed GestureGAN over the state-of-the-art models on the unconstrained hand gesture-to-gesture translation task. In addition, the generated hand gesture images are of high-quality and are photo-realistic, thus allowing them to be used to boost the performance of hand gesture classifiers.
\end{itemize}

\section{Related Work}
\label{sec:relatewprk}

\begin{figure*}[!t] \tiny
	\centering
	\includegraphics[width=0.94\linewidth]{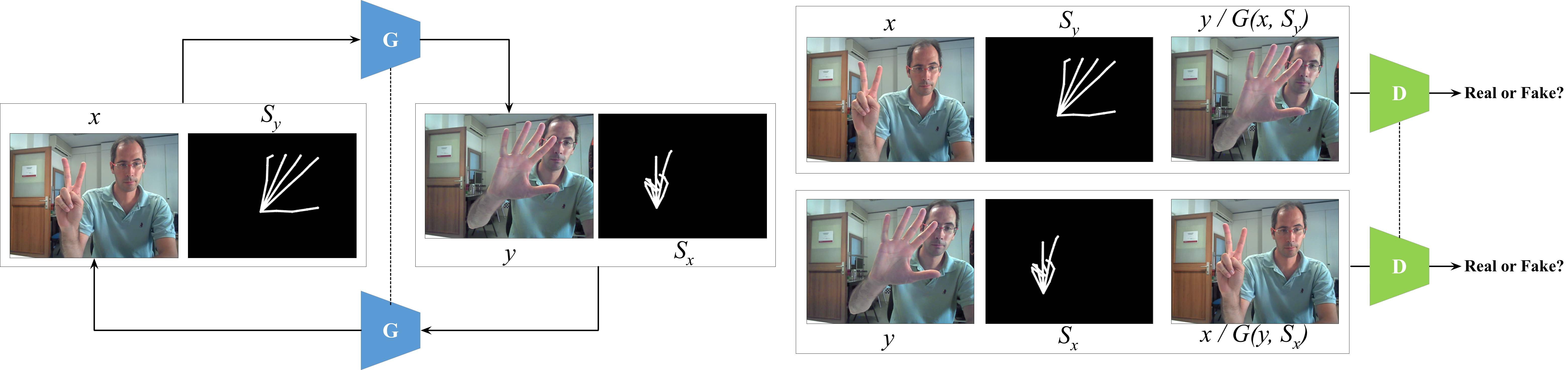}
	\caption{Pipeline of the proposed GestureGAN model. GestureGAN consists of a single generator $G$ and a discriminator $D$, which takes as input a conditional hand image and a target hand skeleton image. 
	}
	\label{fig:framework}
\end{figure*}

\noindent\textbf{Generative Adversarial Network (GAN)} \cite{goodfellow2014generative} is an unsupervised learning method and has been proposed by Goodfellow et al.
Recently, GAN has shown outstanding results in various applications, \textit{e.g.}, image generation \cite{karras2017progressive,berthelot2017began}, image editing \cite{shu2017neural,perarnau2016invertible}, video generation \cite{tulyakov2017mocogan,mathieu2015deep}, texture synthesis \cite{li2016precomputed}, music generation \cite{yang2017midinet} and feature learning \cite{xie2017controllable}.
Recent approaches employ the idea of GAN for conditional image generation, such as image-to-image translation \cite{isola2017image,wang2018high}, text-to-image translation \cite{reed2016generative,han2017stackgan}, image inpainting \cite{li2017generative,dolhansky2017eye},
image blending \cite{wu2017gp},  image super-resolution \cite{ledig2016photo}, as well as the applications of other domains like semantic segmentation \cite{luc2016semantic}, object detection \cite{li2017perceptual,wang2017fast}, human parsing \cite{liu2018cross}, face aging \cite{liu2017face} and 3D vision \cite{park2017transformation,wu2016learning}.
The key point success of GANs in computer vision and graphics is the adversarial loss, which allows the model to generate images that are  indistinguishable from real images, and this is exactly the goal that many computer vision and graphics tasks aim to optimize.

\noindent\textbf{Image-to-Image Translation} frameworks use input-output data to learn a parametric mapping between inputs and outputs, \textit{e.g.}, Isola et al. \cite{isola2017image} build the pix2pix model, which uses a conditional GAN to learn a translation function from input to output image domains.
Taigman et al. \cite{taigman2016unsupervised} propose the Domain Transfer Network (DTN) which learns a generative function between one domain and another domain.
Zhu et al. \cite{zhu2017unpaired} introduce the CycleGAN framework, which achieves unpaired image-to-image translation using the cycle-consistency loss.
Moreover, Zhu et al. \cite{zhu2017toward} present the BicycleGAN model based on CycleGAN \cite{zhu2017unpaired} and pix2pix \cite{isola2017image}, which targets multi-modal image-to-image translation.

However, existing image-to-image translation models are inefficient and ineffective.
For example, with $m$ image domains, CycleGAN~\cite{zhu2017unpaired}, DiscoGAN \cite{kim2017learning}, and DualGAN~\cite{yi2017dualgan} need to train $2C_m^2{=}m(m{-}1){=}\Theta(m^2)$ generators and discriminators, while pix2pix \cite{isola2017image} and BicycleGAN \cite{zhu2017toward} have to train $A_m^2{=}m(m{-}1){=}\Theta(m^2)$ generator/discriminator pairs.
Recently, Anoosheh et al. proposed ComboGAN \cite{anoosheh2017combogan}, which only need to train $m$ generator/discriminator pairs for $m$ different image domains, having a complexity of $\Theta(m)$.
Additionally, Choi et al. \cite{choi2017stargan} propose StarGAN, in which a single generator and discriminator can perform unpaired image-to-image translations for multiple domains.
Although the computational complexity of StarGAN is $\Theta(1)$, this model has only been validated on the face attributes modification task with clear background and face cropping.
More importantly, for some specific image-to-image translation tasks  such as hand gesture-to-gesture translation task, 
the image domains could be arbitrary large, \emph{e.g.}, gesture in the wild can have arbitrary poses, sizes, appearances, locations and self-occlusions.
The aforementioned approaches are not effective for solving these specific situations.

\noindent\textbf{Keypoint/Skeleton Guided Image-to-Image Translation.}
To fix these limitations, several recent works have been proposed to generate person, bird or face images based on object keypoints \cite{reed2016learning,ma2017pose} or human skeleton \cite{yan2017skeleton,siarohin2017deformable}.
For instance,
Di et al.~\cite{di2017gp} propose the Gender Preserving Generative Adversarial Network (GPGAN) to synthesize faces based on facial landmarks.
Reed et al.~\cite{reed2016learning} propose the Generative Adversarial What-Where Network (GAWWN), which generates birds conditioned on both text descriptions and object location.
Ma et al. propose the Pose Guided Person Generation Network (PG$^2$) \cite{ma2017pose} and a two-stage reconstruction pipeline \cite{ma2017disentangled}, which achieve person-to-person image translation using a conditional image and a target pose image (note that in these two models images are pre-cropped).
Reed et al. \cite{reed2016generating} present an extension of Pixel Convolutional Neural Networks (PixelCNN) to generate images parts based on keypoints and text description.
Sun et al. \cite{sun2017natural} propose a two-stage framework to  perform head inpainting conditioned on the generated facial landmark in the first stage.
Korshunova et al. \cite{korshunova2016fast} use facial keypoints to define the affine transformations of the alignment and realignment steps for face swap. 
Wei et al.~\cite{wei2017every} propose a Conditional MultiMode Network (CMM-Net) for landmark-guided  smile generation.
Qiao et al.~\cite{qiao2018geometry} present the Geometry-Contrastive Generative Adversarial Network (GCGAN) to generate facial expressions conditioned on geometry information of facial landmarks. 
Song et al. \cite{song2017geometry} propose the Geometry-Guided Generative Adversarial Network (G2GAN) for facial expression synthesis guided by fiducial points.
Yan et al. \cite{yan2017skeleton} propose a method to generate human motion sequence with simple background using CGAN and human skeleton information.
Siarohin et al. \cite{siarohin2017deformable} introduce PoseGAN for pose-based human image generation using human skeleton.

The typical problem with the aforementioned generative models is that they suffer from ``channels pollution'' and thus
they tend to generate blurry results with artifacts.
To handle this problem, we propose GestureGAN, which allows generating high-quality hand gesture image with arbitrary poses, sizes and locations in the wild.

\section{Model Description}
\label{sec:method}

\subsection{GestureGAN Objective}

The goal of vanilla GAN is to train a generator $G$ which learns the mapping between random noise $z$ and image $y$.
The mapping function of GAN, $G(z)\mapsto y$ is learned via the following objective function,
\begin{equation}\small
\mathcal{L}_{GAN}(G, D) = 
\mathbb{E}_y \left[ \log D(y) \right] +
\mathbb{E}_{z} \left[\log (1{-} D(G(z))) \right].
\end{equation}
Conditional GANs learn the mapping $G(x, z)\mapsto y$, where  $x$ is the input conditional image.
Generator $G$ is trained to generate image $\widehat{y}$ that cannot be distinguished from ``real'' image $y$ by an adversarially trained discriminator $D$, while the discriminator $D$ is trained as well as possible to detect the ``fake'' images generated by the generator $G$.  
The objective function of the conditional GAN is defined as follows,
\begin{equation}\small
\begin{aligned}
& \mathcal{L}_{cGAN}(G, D) = 
\mathbb{E}_{x, y} \left[ \log D(x, y) \right] + 
\mathbb{E}_{x, z} \left[\log (1 - D(x, G(x, z))) \right],
\end{aligned}
\label{eqn:conditonalgan}
\end{equation}
where generator $G$ tries to minimize this objective while the discriminator $D$ tries to maximize it. 
Thus, the solution is $G^*=\arg \min\limits_G \max\limits_D \mathcal{L}_{cGAN}(G, D).$
In this paper, we try to learn two mappings through one generator.
The framework of the proposed GestureGAN is shown in Figure \ref{fig:framework}.

\noindent \textbf{Adversarial Loss.} In order to learn the  gesture-to-gesture mapping, we employ the hand skeleton information explicitly.
We exploit OpenPose \cite{simon2017hand} to detect 21 hand keypoints denoted as $(p_i, q_i \vert i=1,2,\cdots,21)$, where $p_i$ and $q_i$ represent pixel coordinates of keypoints.
For each keypoint $(p_i, q_i)$, $c_i {\in} [0,1]$ represents the confidence that the keypoint is correctly localized.
Thus, the adversarial losses of the two mappings $G([x, K_y], z_1)\mapsto y$  and $G([y, K_x], z_2)\mapsto x$ are defined respectively, as follows:
\begin{equation}\small
\begin{aligned}
\mathcal{L}_{K_y}(G, D, K_y) = &  \mathbb{E}_{[x, K_y], y} \left[\log D([x, K_y], y) \right] +  \\
& \mathbb{E}_{[x,K_y], z_1} \left[\log (1 - D([x, K_y], G([x, K_y], z_1))) \right],
\end{aligned}
\label{eqn:keypointgan}
\end{equation}

\begin{equation}\small
\begin{aligned}
\mathcal{L}_{K_x}(G, D, K_x) =  & \mathbb{E}_{[y, K_x], x} \left[\log D([y, K_x], x) \right] +  \\
& \mathbb{E}_{[y,K_x], z_2} \left[\log (1 - D([y, K_x], G([y, K_x], z_2))) \right],
\end{aligned}
\label{eqn:keypointgan2}
\end{equation}
where $K_y$ and $K_x$ are the hand keypoints of image $y$ and $x$ respectively; 
$[\cdot,\cdot]$ represents the concatenate operation. 
$K_y$ and $K_x$ are defined by setting the pixels around the corresponding keypoint $(p_i, q_i)$ to 1 (white) with the radius of 4 and 0 (black) elsewhere.
In other words, each keypoint is actually represented with pixels in a circle with a radius of 4. 
Therefore, the total adversarial loss based on hand keypoint can be defined as,
\begin{equation}\small
\begin{aligned}
\mathcal{L}_K(G, D, K_x, K_y) = \mathcal{L}_{K_y}(G, D, K_y) + \mathcal{L}_{K_x}(G, D, K_x).
\end{aligned}
\label{eqn:adv}
\end{equation}

In addition, to explore the influence of the confidence score $c_i$ to the generated image, we define a confidence keypoint image $\widehat{K}$ in which the pixels around the corresponding keypoint $(p_i, q_i)$ are set to $c_i$ in a radius of 4 pixels and 0 (black) elsewhere.
Thus, Equation \ref{eqn:adv} can be expressed as $\mathcal{L}_{\widehat{K}}(G, D, \widehat{K_x}, \widehat{K_y}) = \mathcal{L}_{\widehat{K_y}}(G, D, \widehat{K_y}) + \mathcal{L}_{\widehat{K_x}}(G, D, \widehat{K_x})$.

Moreover, following OpenPose \cite{simon2017hand}, we connect the 21 keypoints (hand joints) to obtain the hand skeleton, denoted as $S_x$ and $S_y$.
The hand skeleton image visually contains richer hand structure information than the hand keypoint image.
Next, the adversarial loss based on hand skeleton can be derived from Equation~\ref{eqn:adv}, \textit{i.e.}, $\mathcal{L}_S(G, D, S_x, S_y) = \mathcal{L}_{S_y}(G, D, S_y) + \mathcal{L}_{S_x}(G, D, S_x) $.
In hand skeleton image $S_x$ and $S_y$, the hand joints are connected by the lines with the width of 4 and with white color.
Next, corresponding to the confidence keypoint image, we have also defined an adversarial loss using confidence hand skeleton as $\mathcal{L}_{\widehat{S}}(G, D, \widehat{S_x}, \widehat{S_y}) = \mathcal{L}_{\widehat{S_y}}(G, D, \widehat{S_y}) + \mathcal{L}_{\widehat{S_x}}(G, D, \widehat{S_x}) $,
where the line connections in $\widehat{S_x}$ and $\widehat{S_y}$ are filled with the confidence score of later point, \textit{e.g.}, if the hand skeleton connects points 1 and 2, thus this line connection is filled with the confidence of point 2, \textit{i.e.}, $c_2$ with the width of 4.

\noindent \textbf{Improved Pixel Loss.} Previous work indicated that mixing the adversarial loss with a traditional loss such as $L_1$ loss \cite{isola2017image} or $L_2$ loss \cite{pathak2016context} between the generated image and the ground truth image improves the quality of generated images.
The definition of $L_1$ and $L_2$ losses are:
\begin{equation}\small
\begin{aligned}
\mathcal{L}_{L_{\{1,2\}}}(G, S_x, S_y) = &\mathbb{E}_{[x, S_y], y, z_1} \left[ \lVert y - G([x, S_y], z_1) \lVert_{\{1,2\}} \right] + \\
&\mathbb{E}_{[y, S_x], x, z_2} \left[ \lVert x - G([y, S_x], z_2) \lVert_{\{1,2\}} \right].
\end{aligned}
\label{equ:l2}
\end{equation}

\begin{figure}[!t] \tiny
	\centering
	\includegraphics[width=0.95\linewidth]{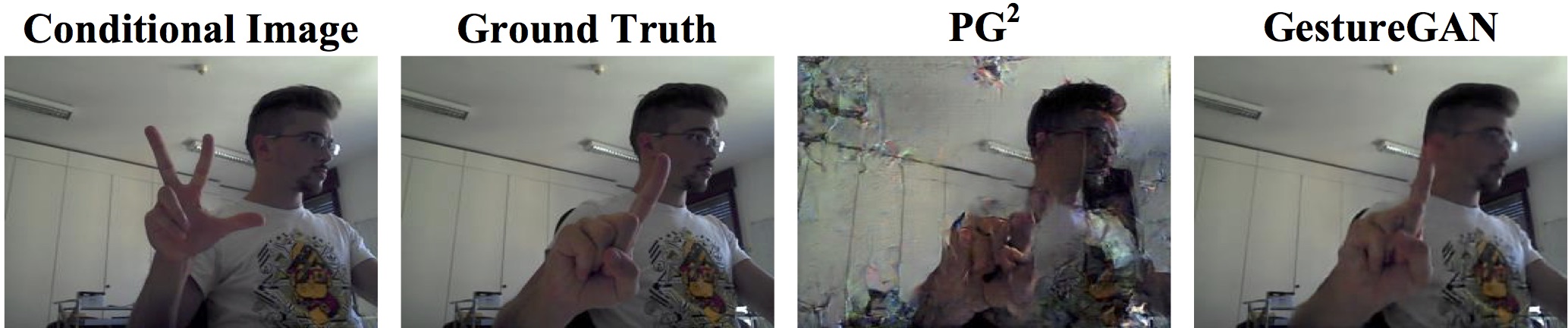}
	\caption{Illustration of the ``channel pollution'' issue on different methods.}
	\label{fig:first_figure}
	\vspace{-0.3cm}
\end{figure}

However, we observe that the existing image-to-image translation models such as PG$^2$ \cite{ma2017pose} cannot retain the holistic color of the input images.
An example is shown in Figure~\ref{fig:first_figure}, where PG$^2$ is affected by the pollution issue and produces more unrealistic regions. 
Therefore, to remedy this limitation we introduce a novel channel-wise color loss.
Traditional generative models convert a whole image to another, which leads to the ``channel pollution'' problem.
However, the color loss treats $r$, $g$ and $b$ channels independently and generates only one channel each time, and then these three channels are combined to produce the final image.
Intuitively, since the generation of a three-channel image space is much more complex than the generation of a single-channel image space, leading to higher possibility of artifacts, we independently generate each channel.
The objective of $r$, $g$ and $b$ channel losses can be defined as follows,

\vspace{-0.3cm}
\begin{equation} \small
\begin{aligned}
\mathcal{L}_{Color^{c}_{\{1,2\}}}(G, S_x, S_y) = & 
\mathbb{E}_{[x^{c}, S_y], y^{c}, z_1} \left[ \lVert y^{c} - G([x^{c}, S_y], z_1) \lVert_{\{1,2\}} \right] + \\
& \mathbb{E}_{[y^{c}, S_x], x^{c}, z_2} \left[ \lVert x^{c} -	 G([y^{c}, S_x], z_2) \lVert_{\{1,2\}} \right],
\end{aligned}
\label{eqn:color_rgb}
\end{equation}
where $c{\in}\{r,g,b\}$, $x^r$, $x^g$ and $x^b$ denote the $r$, $g$ and $b$ channels of image $x$ respectively and similar to $y^r$, $y^g$ and $y^b$, $\lVert\cdotp \lVert_1$ and $\lVert\cdotp \lVert_2$ represent $L_1$ and $L_2$ distance losses.
Thus, the color $L_1$ and $L_2$ losses can be expressed as, 
\begin{equation}\small
\begin{aligned}
\mathcal{L}_{Color_{\{1,2\}}}(G, S_x, S_y) {=} \mathcal{L}_{Color^{r}_{\{1,2\}}} {+} \mathcal{L}_{Color^{g}_{\{1,2\}}} {+} \mathcal{L}_{Color^{b}_{\{1,2\}}}.
\end{aligned}
\label{eqn:color2}
\end{equation}

When back-propagating the gradients of the $L_1$ loss, the partial derivatives are constants, \emph{i.e.}, $\pm 1$. 
Therefore, the error from other channels will not influence the current one as the derivative is a constant.
However, for the original $L_2$ loss, the derivative is not a fixed constant. Actually, the derivative for the variables in one channel is always influenced by the errors from other channels. 
We have listed the gradients of red channel of Equations \ref{equ:l2} and \ref{eqn:color2}. Let $\widehat{y}$ represent the generated target image $G([x, S_y], z_1)$, we have,
\vspace{-0.2cm}
\begin{equation} \small
\begin{aligned}
& \frac{\partial}{\partial \widehat{y}_r^{i_o,j_o}} \mathcal{L}_{L_2}(G, S_x, S_y)  \\
= &  \frac{\partial}{\partial \widehat{y}_r^{i_o,j_o}} \sqrt{\sum_{i,j}(y_r^{i,j}-\widehat{y}_r^{i,j})^2 {+} \sum_{i,j}(y_g^{i,j}-\widehat{y}_g^{i,j})^2 + \sum_{i,j}(y_b^{i,j}-\widehat{y}_b^{i,j})^2} \\
= &  \frac{y_r^{i_o,j_o}-\widehat{y}_r^{i_o,j_o}}{\sqrt{\sum\limits_{i,j}(y_r^{i,j}-\widehat{y}_r^{i,j})^2 {+} \sum\limits_{i,j} (y_g^{i,j}-\widehat{y}_g^{i,j})^2 {+} \sum\limits_{i,j}(y_b^{i,j}-\widehat{y}_b^{i,j})^2}}. 
\end{aligned}
\label{eqn:gra_l2}
\end{equation}

\begin{equation} \small
\begin{aligned}
& \frac{\partial}{\partial \widehat{y}_r^{i_o,j_o}} \mathcal{L}_{Color_{2}}(G, S_x, S_y) \\ 
= &  \frac{\partial}{\partial \widehat{y}_r^{i_o,j_o}}\left( \sqrt{\sum_{i,j}(y_r^{i,j} {-} \widehat{y}_r^{i,j})^2} {+} \sqrt{\sum_{i,j}(y_g^{i,j} {-} \widehat{y}_g^{i,j})^2} {+} \sqrt{\sum_{i,j}(y_b^{i,j} {-} \widehat{y}_b^{i,j})^2} \right) \\ 
= & \frac{y_r^{i_o,j_o}-\widehat{y}_r^{i_o,j_o}}{\sqrt{\sum\limits_{i,j}(y_r^{i,j}-\widehat{y}_r^{i,j})^2}}. 
\end{aligned}
\label{eqn:gra_colorl2}
\end{equation}
Therefore, we can calculate the gradient of original $L_2$ and color $L_2$ losses, 
\begin{equation}\small
\begin{aligned}
\bigtriangledown \mathcal{L}_{L_2}(G, S_y) = &\frac{\partial}{\partial \widehat{y}_r^{i_o,j_o}}\mathcal{L}_{L_2}(G, S_y) + \\
& \frac{\partial}{\partial \widehat{y}_g^{i_o,j_o}}\mathcal{L}_{L_2}(G, S_y) + \frac{\partial}{\partial \widehat{y}_b^{i_o,j_o}}\mathcal{L}_{L_2}(G, S_y).
\end{aligned}
\end{equation}
\vspace{-0.2cm}
\begin{equation}\small
\begin{aligned}
\bigtriangledown \mathcal{L}_{Color_{2}}(G, S_y) = & \frac{\partial}{\partial \widehat{y}_r^{i_o,j_o}} \mathcal{L}_{Color_{2}}(G, S_y) + \\
& \frac{\partial}{\partial \widehat{y}_g^{i_o,j_o}} \mathcal{L}_{Color_{2}}(G, S_y) +
\frac{\partial}{\partial \widehat{y}_b^{i_o,j_o}} \mathcal{L}_{Color_{2}}(G, S_y).
\end{aligned}
\end{equation}

Clearly, in Equation \ref{eqn:gra_l2} the red channel in original $L_2$ loss is polluted by green and blue channels. As a consequence, the error from other channels will also influence the red channel. 
On the contrary, if we compute the loss for each channel independently, we can avoid such influence as shown in Equation \ref{eqn:gra_colorl2}.

\noindent \textbf{Cycle-Consistency Loss.}
It is worth noting that the CycleGAN~\cite{zhu2017unpaired} is different from pix2pix framework \cite{isola2017image} as the training data in CycleGAN is unpaired. 
The CycleGAN introduces the cycle-consistency loss to enforce forward-backward consistency.
The cycle-consistency loss can be regarded as ``pseudo'' pairs of training data even though we do not have the corresponding data in the target domain which corresponds to the input data from the source domain.
However, in this paper we introduce the cycle-consistency loss for the paired image-to-image translation task.
The cycle loss ensures the consistency between source images and the reconstructed image, and it can be expressed as,
\begin{equation} \small
\begin{aligned}
\mathcal{L}_{cyc}(G, S_x, S_y) = &  \mathbb{E}_{x, y, S_x, S_y, z_1, z_2} \left[ \left|\left| x - G(G([x, S_y], z_1), S_x, z_2) \right|\right|_1 \right] + \\
& \mathbb{E}_{x, y, S_x, S_y, z_1, z_2} \left[ \left|\left| y - G(G([y, S_x], z_2), S_y, z_1) \right|\right|_1 \right].
\end{aligned}
\label{eqn:con}
\end{equation}
Similar to StarGAN \cite{choi2017stargan} we use the same generator $G$ two times, with the first time to convert an original image into the target one, then to recover the original image from the generated image.

\noindent \textbf{Identity Preserving Loss.} To preserve the person identity after image synthesis, we propose the identity preserving loss, which can be expressed as follows,
\begin{equation}\small
\begin{aligned}
\mathcal{L}_{identity}(G,S_x, S_y) = & \mathbb{E}_{x, S_y, z_1} \left[ \left|\left| F(y) - F(G([x, S_y], z_1)) \right|\right|_1 \right]+\\
&  \mathbb{E}_{y, S_x, z_2} \left[ \left|\left| F(x) - F(G([y, S_x], z_2)) \right|\right|_1 \right],
\end{aligned}
\label{eqn:preserve}
\end{equation}
where $F$ is a feature extractor. 
The feature extractor employs a VGG network \cite{simonyan2014very} originally pretrained for face recognition.
We minimize the difference between the feature maps which are generated from the real and the generated images via the pretrained CNN for identity preservation. 

\noindent \textbf{Overall Loss.} The final objective of the proposed GestureGAN is,
\begin{equation}\small
\begin{aligned}
\mathcal{L} = & \mathcal{L}_{S}(G, D, S_x, S_y)  + \lambda_1 \mathcal{L}_{Color_{\{1, 2\}}}(G, S_x, S_y) + \\ 
& \lambda_2 \mathcal{L}_{cyc}(G, S_x, S_y) + \lambda_3 \mathcal{L}_{identity}(G,S_x, S_y)	,
\end{aligned}
\label{eqn:overallloss}
\end{equation}
where, $\lambda_1$ , $\lambda_2$ and $\lambda_3$ are three hyper-parameters controlling the relative importance of these four losses.
In our experiments, we follow the same setup of pix2pix \cite{isola2017image}.
Instead of using the random noise vector~$z$, we provide noise only in the form of dropout in generator~$G$.

\subsection{Network Architecture}

\noindent\textbf{Generator.}
We adopt the ``U-shaped'' network \cite{isola2017image} as our generator.
U-net has skip connections, which concatenate all channels at layer $l$ with those at layer $n{-}l$, where $n$ is the total number of layers.

\noindent\textbf{Discriminator.}
We employ PatchGAN  \cite{isola2017image}  as our discriminator architecture.
The goal of PatchGAN  is to classify each small patch in an image as real or fake.
We run PatchGAN convolutationally across an image, then average all results to calculate the ultimate output of discriminator $D$.

\subsection{Optimization}

We follow the standard optimization method from \cite{goodfellow2014generative} to optimize the proposed GestureGAN, \textit{i.e.}, we alternate between one gradient descent step on discriminator $D$, and one step on generator $G$.
In addition, as suggested in the original GAN paper \cite{goodfellow2014generative}, we train to maximize $\log D([x, S_y], \widehat{y})$ rather than to minimize $\log (1 - D([x, S_y], \widehat{y}))$.
Moreover, in order to slow down the rate of $D$ relative to $G$ we divide the objective by 2 while optimizing $D$,
\vspace{-0.1cm}
\begin{equation} \small
\begin{aligned}
\mathcal{L}(D) = & 
\frac{1}{2}\left[\mathcal{L}_{bce}(D([x, S_y], y), 1) + \mathcal{L}_{bce}(D([x,S_y], G([x,S_y],z_1)), 0)\right] \\ + 
& \frac{1}{2}\left[\mathcal{L}_{bce}(D([y, S_x], x), 1) + \mathcal{L}_{bce}(D([y,S_x], G([y,S_x],z_2)), 0) \right],
\end{aligned}
\label{eqn:dloss}
\end{equation}
where $\mathcal{L}_{bce}$ denotes the Binary Cross Entropy loss function.
We also employ dual discriminators as in Xu et al. \cite{xu2017face}, Nguyen et al. \cite{nguyen2017dual} and CycleGAN \cite{zhu2017unpaired}, which have demonstrated that they improve the ability of discriminator to generate more photo-realistic images.
Thus Equations \ref{eqn:dloss} is modified as: 
\vspace{-0.1cm}
\begin{equation}\small
\begin{aligned}
& \mathcal{L}(D_1, D_2) =  \\ 
&\frac{1}{2}\left[\mathcal{L}_{bce}(D_1([x, S_y], y), 1) + \mathcal{L}_{bce}(D_1([x,S_y], G([x,S_y],z_1)), 0)\right] + \\
& \frac{1}{2}\left[\mathcal{L}_{bce}(D_2([y, S_x], x), 1) + \mathcal{L}_{bce}(D_2([y,S_x], G([y,S_x],z_2)), 0) \right].
\end{aligned}
\label{equ:double_d1}
\end{equation}
We employ the minibatch SGD algorithm and apply Adam optimizer~\cite{kingma2014adam} as solver. The momentum terms $\beta_1$ and  $\beta_2$ of Adam are 0.5 and 0.999, respectively.  
The initial learning rate for Adam is 0.0002.

\section{Experiments}
\label{sec:experiment}

\subsection{Experimental Setup}
\noindent \textbf{Datasets.}
We evaluate the proposed GestureGAN on two public hand gesture datasets: NTU Hand Digit \cite{ren2013robust} and Creative Senz3D \cite{memo2016head}, which include different hand gestures.
NTU Hand Digit dataset \cite{ren2013robust} contains 10 hand gestures (\textit{e.g.}, decimal digits from 0 to 9) color images
and depth maps collected with a Kinect sensor under cluttered background.
The total images in this dataset are 10 gestures $\times$ 10 subjects ${\times}$ 10 times = 1000 images.
All images are in $640 {\times} 480$ resolution.
In our experiment, we only use the RGB images.
We randomly select 84,636 pairs, each of which is comprised of two images of the same person but different gestures.
9,600 pairs are randomly selected for the testing subset and the rest of 75,036 pairs as the training set.
Creative Senz3D dataset \cite{memo2016head} includes static hand gestures performed by 4 people, each performing 11 different gestures repeated 30 times each in
the front of a Creative Senz3D camera.
The overall number of images of this dataset is 4 subjects $\times$ 11 gestures $\times$ 30 times = 1320.
All images are in resolution $640 {\times} 480$.
In our experiment, we only use the RGB images.
We randomly select 12,800 pairs and 135,504 pairs as the testing and training set, each pair being composed of two images of the same person but different gestures.

\noindent\textbf{Implementation Details.}
For both datasets, we do left-right flip for data augmentation and random crops are disabled in this experiment as was done in PG$^2$ \cite{ma2017pose}.
For the embedding method, skeleton images are fed into an independent encoder similar to PG$^2$ \cite{ma2017pose},  then we extract the fully connected layer feature vector to concatenate it with the image embedding at the bottleneck fully connected layer.
For optimization, models are trained with a mini-batch size of 8 for 20 epochs on both datasets.
Hyper-parameters are set empirically with $\lambda_1{=}100$,  $\lambda_2{=}10$.
$\lambda_3$ is 0.1 in the beginning and is gradually increased to 0.5.
At inference time, we follow the same settings of PG$^2$~\cite{ma2017pose}, Ma et al. \cite{ma2017disentangled} and PoseGAN \cite{siarohin2017deformable} to randomly select the target keypoint or skeleton.
GestureGAN is implemented using the public deep learning framework PyTorch.
To speed up the training and testing processes, we use a Nvidia TITAN Xp GPU with 12G memory.

\begin{figure*}[!t] \tiny
	\centering
	\includegraphics[width=0.95\linewidth]{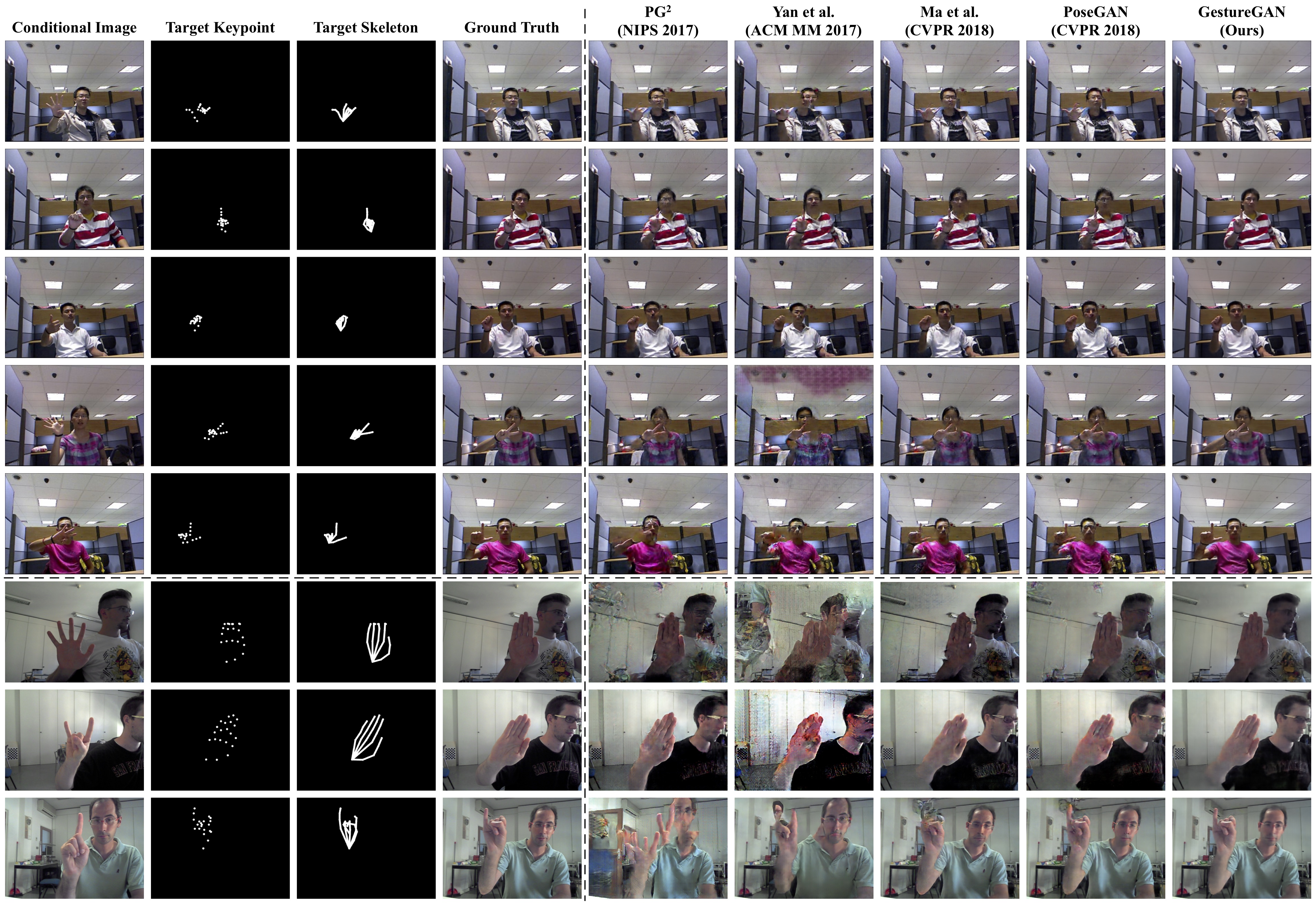}
	\caption{Qualitative comparison with PG$^2$ \cite{ma2017pose}, Ma et al. \cite{ma2017disentangled}, Yan et al. \cite{yan2017skeleton} and PoseGAN \cite{siarohin2017deformable} on the NTU Hand Digit (Top) and the Senz3D (Bottom) datasets. Zoom in for details.}
	\label{fig:comparsion}
\end{figure*}

\begin{table*}[!t] \small
	\centering
	\vspace{-0.2cm}
	\caption{Quantitative results of different models on the NTU Hand Digit and Senz3D datasets. For PSNR and IS measures, higher is better. For MSE evaluation, lower is better.}
	\begin{tabular}{l|c|c|c|c|c|c} \toprule
		\multirow{2}{*}{Model} & \multicolumn{3}{c|}{NTU Hand Digit \cite{liu2016deepfashion}}& \multicolumn{3}{c}{Senz3D \cite{memo2016head}}   \\ \cline{2-7}
		& MSE               & PSNR             & IS              & MSE               & PSNR             & IS     \\ \midrule			
		PG$^2$ \cite{ma2017pose}  (NIPS 2017)                                        & 116.1049          & 28.2403          & 2.4152          & 199.4384          & 26.5138          & 3.3699 \\ \hline
		Yan et al. \cite{yan2017skeleton} (ACM MM 2017)                              & 118.1239          & 28.0185          & 2.4919          & 175.8647          & 26.9545          & 3.3285 \\ \hline
		Ma et al. \cite{ma2017disentangled} (CVPR 2018)                              & 113.7809          & 30.6487          & 2.4547          & 183.6457          & 26.9451          & 3.3874 \\ \hline 
		PoseGAN \cite{siarohin2017deformable} (CVPR 2018)                            & 113.6487          & 29.5471          & 2.4017          & 176.3481          & 27.3014          & 3.2147 \\ \hline \hline
		GestureGAN (Ours)                                                            & \textbf{105.7286} & \textbf{32.6091} & \textbf{2.5532 }         & \textbf{169.9219} & \textbf{27.9749}          & \textbf{3.4107} \\ \bottomrule		
	\end{tabular}
	\label{tab:baseline}
	\vspace{-0.2cm}
\end{table*}

\begin{figure*}[!t] \tiny
	\centering
	\includegraphics[width=0.95\linewidth]{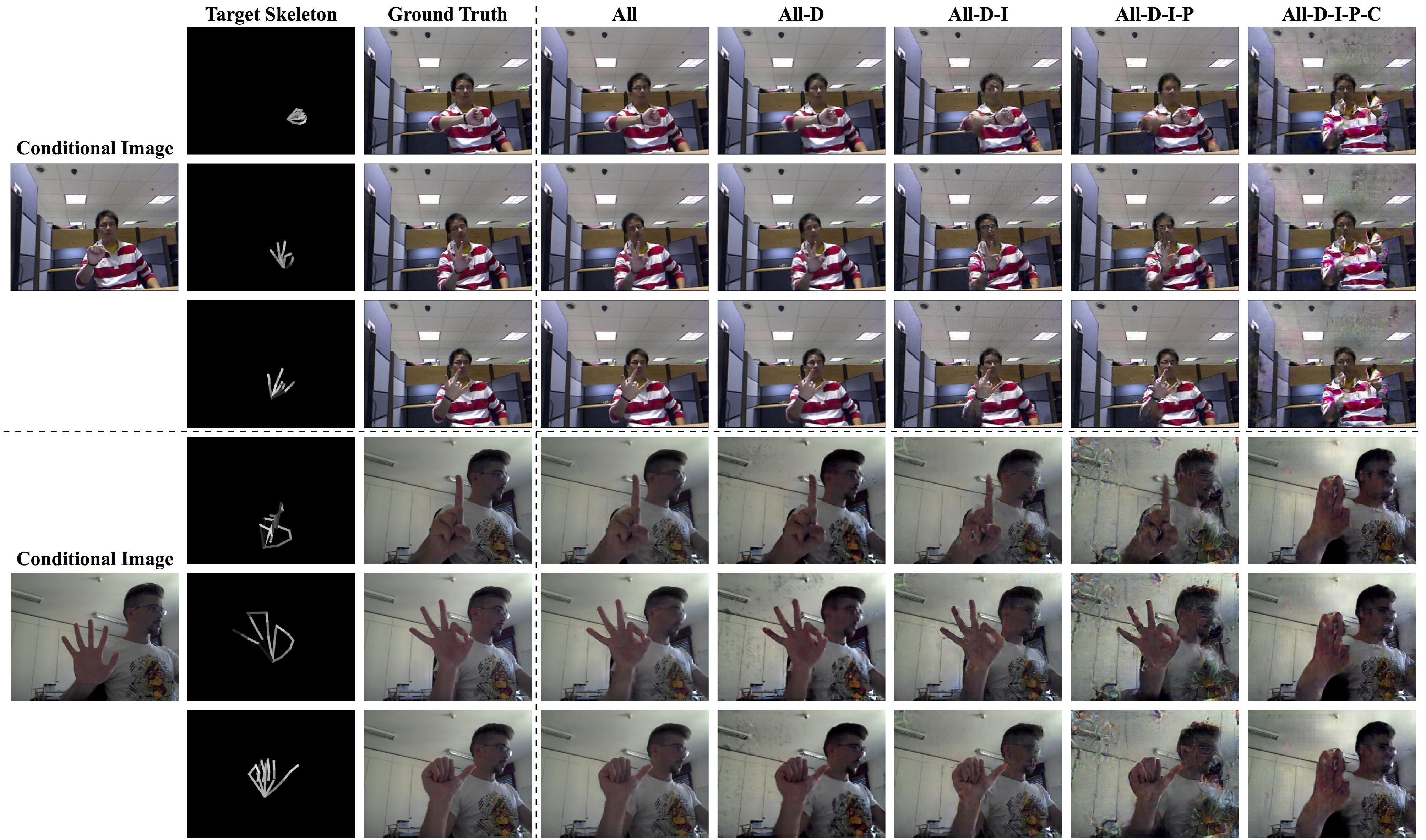}
	\caption{Qualitative comparison using different components of GestureGAN on the NTU Hand Digit (Top) and the Senz3D (Bottom) datasets.  All: full version of GestureGAN, D: Dual discriminators strategy, I: Identity preserving loss, P: Color loss, C: Cycle-consistency loss. ``-'' means removing. Zoom in for details.}
	\label{fig:loss}
\end{figure*}

\noindent\textbf{Evaluation Metrics.}
Mean Squared Error (MSE), Peak Signal-to-Noise Ratio (PSNR), Inception Score (IS) \cite{salimans2016improved}, Fr\'echet Inception Distance (FID) \cite{heusel2017gans} and the proposed Fr\'echet ResNet Distance (FRD) are employed to evaluate the quality of generated images.
FRD approach provides an alternative method to quantify the quality of synthesis and is similar to the Fr\'echet Inception Distance (FID) proposed by \cite{heusel2017gans}.
FID is a measure of similarity between two datasets of images. 
The authors in \cite{heusel2017gans} have shown that the FID is more robust to noise than IS \cite{salimans2016improved} and  correlates well with the human judgment of visual quality \cite{heusel2017gans}.
To calculate FID \cite{heusel2017gans} between two image domains $x$ and $y$, they first embed both into a feature space $\mathcal{F}$ given by an Inception model.
Then viewing the feature space as a continuous multivariate Gaussian as suggested in \cite{heusel2017gans}, the Fr\'echet distance between the two Gaussians to quantify the quality of the data and the definition of FID can be expressed as:
\begin{equation} \small
\begin{aligned}
\mathrm{FID}(x, y) = \lVert \mu_x - \mu_y \lVert_2^2 + \mathrm{Tr}(\textstyle{\sum_x + \sum_y} - 2 (\textstyle{\sum_x \sum_y})^{\frac{1}{2}}),
\end{aligned}
\label{eqn:fid}
\end{equation}
where $( \mu_x, \textstyle{\sum_x})$ and $(\mu_y, \textstyle{\sum_y})$ are the mean and the covariance of the data distribution and model distribution, respectfully.

Unlike FID, which regards the datasets $x$ and $y$ as a  whole, the proposed FRD  is inspired from feature matching methods \cite{tang2016novel,zheng2014packing,zheng2014bayes}, and separately calculates the Fr\'echet distance between generated images and real images from the semantical level.
In this way, images from two domains do not affect each other when computing the Fr\'echet distance.  
Moreover, for FID the number of samples should be greater than the dimension of the coding layer, while the proposed FRD does not have this limitation. 
We denote $x_i$ and $y_i$ as images in the  $x$ and $y$ domains, respectively.
For calculating FRD, we first embed both images $x_i$ and $y_i$ into a feature space $\mathcal{F}$ with $1000{\times}1$ dimension given by a ResNet pretrained model \cite{he2016deep}.
We then calculate the Fr\'echet distance between two feature maps $f(x_i)$ and $f(y_i)$.
The Fr\'echet distance $F(f(x_i),f(y_i))$ is defined as the infimum over all reparameterizations $\alpha$ and $\beta$ of $[0, 1]$ of the maximum over all $t \in [0, 1]$ of the distance in $\mathcal{F}$ between $f(x_i)(\alpha(t))$ and $f(y_i)(\beta(t))$, where $\alpha$ and $\beta$ are continuous, non-decreasing surjections of the range $[0,1]$.
The proposed FRD is a measure of similarity between the feature vector of the real image $f(y_i)$  and the feature vector of the generated image $f(x_i)$ by calculating the Fr\'echet distance between them. The Fr\'echet distance is defined as the minimum cord-length sufficient to join a point traveling forward along $f(y_i)$ and one traveling forward along $f(x_i)$, although the rate of travel for each point may not necessarily be uniform. 
Thus, the definition of FRD between two image domain $x$ and $y$ is,
\begin{equation} \small
\begin{aligned}
\mathrm{FRD}(x, y) = \frac{1}{N} \sum_1^N \inf\limits_{\alpha, \beta} \max\limits_{t \in\left[0, 1\right]} \left\lbrace  d \big(f(x_i) (\alpha (t) ), f(y_i) (\beta(t)) \big) \right\rbrace , 
\end{aligned}
\label{eqn:frd}
\end{equation}
where $d$ is the distance function of $\mathcal{F}$, $N$ is the total number of images in $x$ and $y$ domains.

\subsection{Qualitative \& Quantitative Results}
\noindent \textbf{Comparison against Baselines.}
We compare the proposed GestureGAN with the most related four works,
\textit{i.e.}, PG$^2$ \cite{ma2017pose}, Yan et al. \cite{yan2017skeleton}, PoseGAN \cite{siarohin2017deformable} and Ma et al. \cite{ma2017disentangled}.
PG$^2$ \cite{ma2017pose} and Ma et al. \cite{ma2017disentangled} try to generate a person image with different poses based on a conditional person image and a target keypoint image.
Yan et al. \cite{yan2017skeleton} and PoseGAN \cite{siarohin2017deformable} explicitly employ human skeleton information to generate person images.  
Note that Yan et al. \cite{yan2017skeleton} adopt a CGAN to generate motion sequences based on appearance information and skeleton information by exploiting frame level smoothness.
We re-implemented this model to generate a single frame for fair comparison. 
These four methods are paired image-to-image models and comparison results are shown in Figure \ref{fig:comparsion} and Table \ref{tab:baseline}.
As we can see in Figure \ref{fig:comparsion}, GestureGAN produces sharper images with convincing details compared with other baselines.
Moreover, it is obvious that our results in Table \ref{tab:baseline} are consistently much better than baseline methods on both datasets.

\begin{table*}[!t] \small
	\centering
	\vspace{-0.2cm}
	\caption{Ablation study: quantitative results with different components of GestureGAN on the NTU Hand Digit and Senz3D datasets. For PSNR and IS measures, higher is better. For MSE evaluation, lower is better.  All: full version of GestureGAN, D: Dual discriminators strategy, I: Identity preserving loss, P: Color loss, C: Cycle-consistency loss. ``-'' means removing.}
	\begin{tabular}{l|c|c|c|c|c|c} \toprule
		\multirow{2}{*}{Component} & \multicolumn{3}{c|}{NTU Hand Digit \cite{liu2016deepfashion}}& \multicolumn{3}{c}{Senz3D \cite{memo2016head}}   \\ \cline{2-7}
		& MSE      & PSNR    & IS                            & MSE       & PSNR    & IS       \\ \midrule	
		All	                                  & \textbf{105.7286}  & \textbf{32.6091} &2.5532      & \textbf{169.9219} & \textbf{27.9749}          & 3.4107 \\ \hline
		All - D (Dual Discriminators Strategy)         & 118.7830 & 28.0189 & 2.5071                        & 198.0646  & 26.7237 & 3.2740 \\ \hline	
		All - D - I (Identity Preserving Loss)           & 198.7054 & 25.8474 & 2.5438                        & 1319.3957 & 18.3892 & \textbf{4.0784} \\ \hline
		All - D - I - P (Color Loss)          & 406.1478 & 22.1564 & 2.5842                        & 1745.3214 & 14.6598 & 3.4519 \\ \hline 
		All - D - I - P - C (Cycle-Consistency Loss)      & 707.6053 & 20.2684 & \textbf{2.6114}               & 2064.8428 & 15.5426 & 3.2064  \\ \hline	
	\end{tabular}
	\label{tab:result}
	\vspace{-0.2cm}
\end{table*}

\begin{figure}[!t] \tiny
	\centering
	\setcounter{subfigure}{0}
	\subfigure{	\includegraphics[width=0.4\linewidth]{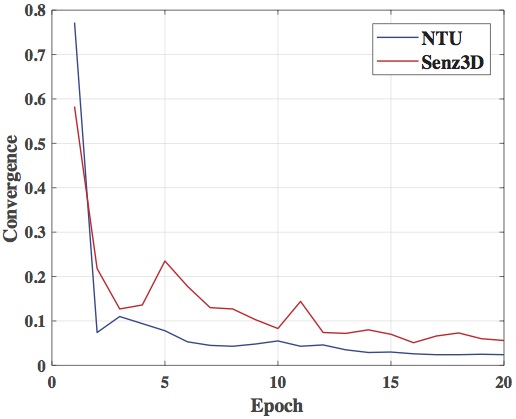}}
	\subfigure{\includegraphics[width=0.4\linewidth]{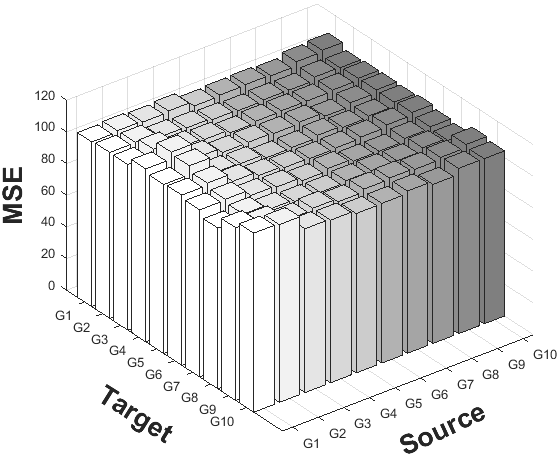}}
	\caption{Convergence loss $\mathcal{L}$ in Equation \ref{eqn:overallloss} (Left) and MSE of different gesture pairs on the NTU dataset (Right).}	
	\label{fig:plot_loss_change}
	\vspace{-0.4cm}
\end{figure}

\noindent \textbf{Generated Results of Each Epoch.}
Figure \ref{fig:plot_loss_change} (left) illustrates the convergence loss $\mathcal{L}$ of the proposed GestureGAN in Equation \ref{eqn:overallloss}. 
Note that the proposed GestureGAN ensures a very fast yet stable convergence.

\noindent \textbf{Analysis of the Model Components.}
In Figure \ref{fig:loss} we conduct ablations studies of our model.
We gradually remove components of the proposed GestureGAN, \textit{i.e.}, Dual Discriminators (D), Identity Loss (I), Color Loss (P) and Cycle-consistency Loss (C).
We find that removing the color loss and the cycle-consistency loss substantially degrades results, meaning that the color loss and the cycle-consistency loss are critical to our results.
In addition, the results without using the identity loss and the dual discriminators slightly degrade performance.
We also provide quantitative results in Table~\ref{tab:result}, and we can see that the full version of GestureGAN  produces more photo-realistic results that other variants on two measurements except IS.
The reason could be that the datasets we used only include human images which do not fit into ImageNet classes \cite{deng2009imagenet}. 
Moreover, PG$^2$ \cite{ma2017pose} and other super-resolution works such as \cite{johnson2016perceptual}  also show the fact that sharper results have a lower quantitative value.

\noindent \textbf{User Study.}
Similar to \cite{ma2017pose,zhu2017unpaired,siarohin2017deformable}, we have also provided a user study. 
We follow the same settings as in \cite{isola2017image} to conduct the Amazon Mechanical Turk (AMT) perceptual studies.
The results of  NTU Hand Digit \cite{liu2016deepfashion} and Senz3D \cite{memo2016head} datasets compared with the baseline models PG$^2$ \protect\cite{ma2017pose}, Ma et al \cite{ma2017disentangled}, Yan et al. \cite{yan2017skeleton} and PoseGAN \cite{siarohin2017deformable} are shown in Table \ref{tab:user}.
Note that the proposed GestureGAN consistently achieves the best performance compared with baselines.  

\noindent \textbf{FID vs. FRD.}
We also compare the performance between FID and the proposed FRD.
The results shown in Table \ref{tab:frd} and we can observe that FRD is more consistent with the human judgment in Table \ref{tab:user} than the FID metric.
Moreover, we observe that the difference in FRD between GestureGAN and the other methods is not as obvious as in the results from the user study in Table \ref{tab:user}.
The reason is that FRD calculates the Fr\'echet distance between the feature maps extracted from the real image and the generated image using CNNs which are trained with semantic labels. Thus, these feature maps are employed to reflect the semantic distance between the images. The semantic distance between the images is not very large considering they are all hands. On the contrary, the user study measures the generation quality from a perceptual level. The difference on the perceptual level is more obvious than on the semantic level, \emph{i.e.}, the generated images with small artifacts show minor difference on the feature level, while being judged with a significant difference from the real images by humans. 

\begin{table}[!t]
	\centering
	\caption{Comparison of AMT perceptual studies (\%) on the NTU Hand Digit and Senz3D datasets.} 
	\resizebox{\linewidth}{!}{%
		\begin{tabular}{l|c|c} \toprule
			Method                                             & NTU Hand Digit \cite{liu2016deepfashion} & Senz3D \cite{memo2016head} \\ \midrule
			PG$^2$ \cite{ma2017pose}  (NIPS 2017)              & 3.5\%                                    & 2.8\%  \\ \hline
			Yan et al. \cite{yan2017skeleton}  (ACM MM 2017)   & 2.6\%                                    & 2.3\%  \\ \hline
			Ma et al. \cite{ma2017disentangled}  (CVPR 2018)   & 7.1\%                                    & 6.9\% \\ \hline
			PoseGAN \cite{siarohin2017deformable} (CVPR 2018)  & 9.3\%                                    & 8.6\%   \\ \hline
			GestureGAN (Ours)                         & \textbf{26.1}\%                                   & \textbf{22.6}\%   \\ \bottomrule		
	\end{tabular}}
	\label{tab:user}
	\vspace{-0.2cm}
\end{table}

\begin{table}[!t] 
	\centering
	\caption{Comparison of FID and the proposed FRD metrics on the NTU Hand Digit and Senz3D datasets. For both FID and FRD, lower is better.}
	\resizebox{\linewidth}{!}{%
		\begin{tabular}{l|c|c|c|c} \toprule
			\multirow{2}{*}{Method} & \multicolumn{2}{c|}{NTU Hand Digit \cite{liu2016deepfashion}}& \multicolumn{2}{c}{Senz3D \cite{memo2016head}}   \\ \cline{2-5}
			& FID & FRD                & FID     & FRD    \\ \midrule			
			PG$^2$ \cite{ma2017pose}  (NIPS 2017)               & 24.2093 & 2.6319             & 31.7333 & 3.0933 \\ \hline
			Yan et al. \cite{yan2017skeleton}  (ACM MM 2017)    & 31.2841 & 2.7453             & 38.1758 & 3.1006 \\ \hline
			Ma et al. \cite{ma2017disentangled}  (CVPR 2018)    & \textbf{6.7661} & 2.6184             & 26.2713 & 3.0846 \\ \hline	
			PoseGAN \cite{siarohin2017deformable} (CVPR 2018)   & 9.6725 & 2.5846             & 24.6712 & 3.0467 \\ \hline		
			GestureGAN (Ours)                                   & 7.5860 & \textbf{2.5223}    & \textbf{18.4595} & \textbf{2.9836} \\  \bottomrule		
	\end{tabular}}
	\label{tab:frd}
	\vspace{-0.3cm}
\end{table}

\noindent \textbf{Data Augmentation.}
The generated images are high-quality and are photo-realistic, and  these images can be used to improve the performance of a hand gesture classifier.
The intuition is that if the generated images are realistic, the classifiers trained on both the real images and the generated images will be able to boost the accuracy of the real images.
In this situation, the generated images work as augmented data.
We employ a pretrained ResNet-50 model \cite{he2016deep} and feed the generated images to fine-tune it. 
For both datasets, we make a split of 70\%/30\% between training and testing sets.
Specifically, the NTU Hand Digit dataset has 700 and 300 images for training and testing set.
For the Senz3D dataset, the numbers of training and testing set are 924 and 396. 
The recognition results for the NTU Hand Digit and the Senz3D datasets are 15\% and 34.34\%, respectively.
The term ``real/real'' in Table~\ref{tab:data_aug} represents the result without data augmentation.
After adding the generated images by different methods to the training set, the performance improves significantly.
Results compared with PG$^2$ \cite{ma2017pose}, Yan et al.~\cite{yan2017skeleton}, Ma et al. \cite{ma2017disentangled} and PoseGAN \cite{siarohin2017deformable}  are shown in Table \ref{tab:data_aug}.
Clearly, GestureGAN achieves the best result compared with baselines.

\begin{table}[!t]
	\centering
	\caption{Comparison of hand gesture recognition accuracy~(\%) on the NTU Hand Digit and Senz3D datasets.} 
	\resizebox{\linewidth}{!}{%
		\begin{tabular}{l|c|c} \toprule
			Method                                                 & NTU Hand Digit \cite{liu2016deepfashion}   & Senz3D \cite{memo2016head}  \\ \midrule
			real/real                                              & 15.000\%          & 34.343\% \\ \hline
			PG$^2$ \cite{ma2017pose}  (NIPS 2017)                  & 93.667\%          & 98.737\%  \\ \hline
			Yan et al. \cite{yan2017skeleton}  (ACM MM 2017)       & 95.333\%          & 99.495\%  \\ \hline
			Ma et al. \cite{ma2017disentangled}  (CVPR 2018)       & 95.864\%          & 99.054\% \\ \hline
			PoseGAN \cite{siarohin2017deformable} (CVPR 2018)      & 96.128\%          & 99.549\% \\ \hline
			GestureGAN (Ours)                             & \textbf{96.667\%} & \textbf{99.747\%}  \\ \bottomrule
	\end{tabular}}
	\label{tab:data_aug}
	\vspace{-0.15cm}
\end{table}

\begin{figure}[!t] \tiny
	\centering
	\includegraphics[width=0.95\linewidth]{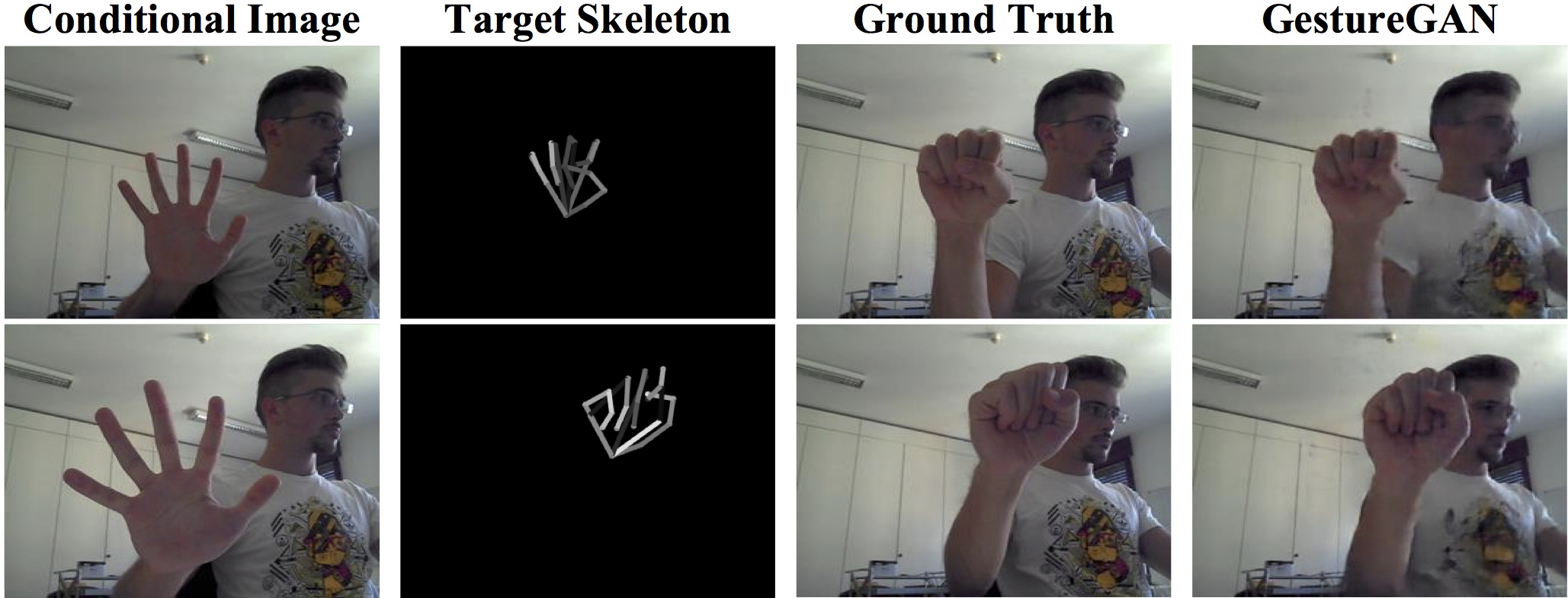}
	\caption{\small{Two samples with different hand sizes and distances.}}
	\label{fig:comparison_size}
	\vspace{-0.3cm}
\end{figure}

\noindent\textbf{Influence of Gesture Size and Distance.} We have also investigated the influence of the gesture size and the distance between source and target gestures. The training samples for the source and the target gesture  from the same person are randomly paired and both gestures have different sizes and distances. Thus, the model is able to learn a robust translation \textit{w.r.t} different hand size and distance. 
We show a qualitative example in Figure~\ref{fig:comparison_size} in which the source images have different sizes and the target gestures have different locations. 
Note that GestureGAN can generate the target gesture from different hand sizes and distances with high quality.

\noindent\textbf{Influence of Gesture Pairs.}
To evaluate the influence of gesture pairs, we searched all the translations between every possible category combinations including the translation within each category.
In Figure \ref{fig:plot_loss_change} (right), we show the MSE for the translation from the source to the target gesture types on NTU dataset.
Note that the MSE for the generation of different gesture pairs has a small variance, showing that the influence of different gesture pairs is very low. This proves that our model is stable. 

\section{Conclusions}
\label{sec:conclusions}

In this paper, we focus on a challenging task of hand gesture-to-gesture translation in the wild.
To this end, we propose a novel Generative Adversarial Network (GAN), \textit{i.e.}, GestureGAN, which can generate  hand gestures with different poses, sizes and locations in the wild.
We also propose two novel losses to learn the mapping from the source gesture to the target gesture, \textit{i.e.}, the color loss and the cycle-consistency loss.
It is worth noting that the proposed color loss handles the ``channel pollution'' problem while back-propagating the gradients, which frequently occurs in the existing generative models.
In addition, we present the Fr\'echet ResNet Distance (FRD) metric to evaluate the quality of generated images.
Experimental results show that GestureGAN achieves state-of-the-art performance.
Lastly, the generated images of GestureGAN are of high-quality and are photo-realistic, and they can thus be used to improve the performance of hand gesture classifiers. 
Future work will focus on designing a GAN model which can handle the situation where the background is total different between source and target gestures.

\section*{Acknowledgements} We want to thank the Nvidia Corporation for the donation of the TITAN Xp GPUs used in this work.

\clearpage
\bibliographystyle{ACM-Reference-Format}
\bibliography{bibliography}


\begin{thebibliography}{00}


\ifx \showCODEN    \undefined \def \showCODEN     #1{\unskip}     \fi
\ifx \showDOI      \undefined \def \showDOI       #1{#1}\fi
\ifx \showISBNx    \undefined \def \showISBNx     #1{\unskip}     \fi
\ifx \showISBNxiii \undefined \def \showISBNxiii  #1{\unskip}     \fi
\ifx \showISSN     \undefined \def \showISSN      #1{\unskip}     \fi
\ifx \showLCCN     \undefined \def \showLCCN      #1{\unskip}     \fi
\ifx \shownote     \undefined \def \shownote      #1{#1}          \fi
\ifx \showarticletitle \undefined \def \showarticletitle #1{#1}   \fi
\ifx \showURL      \undefined \def \showURL       {\relax}        \fi
\providecommand\bibfield[2]{#2}
\providecommand\bibinfo[2]{#2}
\providecommand\natexlab[1]{#1}
\providecommand\showeprint[2][]{arXiv:#2}

\bibitem[\protect\citeauthoryear{Anoosheh, Agustsson, Timofte, and
  Van~Gool}{Anoosheh et~al\mbox{.}}{2018}]%
        {anoosheh2017combogan}
\bibfield{author}{\bibinfo{person}{Asha Anoosheh}, \bibinfo{person}{Eirikur
  Agustsson}, \bibinfo{person}{Radu Timofte}, {and} \bibinfo{person}{Luc
  Van~Gool}.} \bibinfo{year}{2018}\natexlab{}.
\newblock \showarticletitle{ComboGAN: Unrestrained Scalability for Image Domain
  Translation}. In \bibinfo{booktitle}{{\em CVPR Workshops}}.
\newblock


\bibitem[\protect\citeauthoryear{Berthelot, Schumm, and Metz}{Berthelot
  et~al\mbox{.}}{2017}]%
        {berthelot2017began}
\bibfield{author}{\bibinfo{person}{David Berthelot}, \bibinfo{person}{Tom
  Schumm}, {and} \bibinfo{person}{Luke Metz}.} \bibinfo{year}{2017}\natexlab{}.
\newblock \showarticletitle{Began: Boundary equilibrium generative adversarial
  networks}.
\newblock \bibinfo{journal}{{\em arXiv preprint arXiv:1703.10717\/}}
  (\bibinfo{year}{2017}).
\newblock


\bibitem[\protect\citeauthoryear{Choi, Choi, Kim, Ha, Kim, and Choo}{Choi
  et~al\mbox{.}}{2018}]%
        {choi2017stargan}
\bibfield{author}{\bibinfo{person}{Yunjey Choi}, \bibinfo{person}{Minje Choi},
  \bibinfo{person}{Munyoung Kim}, \bibinfo{person}{Jung-Woo Ha},
  \bibinfo{person}{Sunghun Kim}, {and} \bibinfo{person}{Jaegul Choo}.}
  \bibinfo{year}{2018}\natexlab{}.
\newblock \showarticletitle{StarGAN: Unified Generative Adversarial Networks
  for Multi-Domain Image-to-Image Translation}. In \bibinfo{booktitle}{{\em
  CVPR}}.
\newblock


\bibitem[\protect\citeauthoryear{Deng, Dong, Socher, Li, Li, and Fei-Fei}{Deng
  et~al\mbox{.}}{2009}]%
        {deng2009imagenet}
\bibfield{author}{\bibinfo{person}{Jia Deng}, \bibinfo{person}{Wei Dong},
  \bibinfo{person}{Richard Socher}, \bibinfo{person}{Li-Jia Li},
  \bibinfo{person}{Kai Li}, {and} \bibinfo{person}{Li Fei-Fei}.}
  \bibinfo{year}{2009}\natexlab{}.
\newblock \showarticletitle{Imagenet: A large-scale hierarchical image
  database}. In \bibinfo{booktitle}{{\em CVPR}}.
\newblock


\bibitem[\protect\citeauthoryear{Di, Sindagi, and Patel}{Di
  et~al\mbox{.}}{2018}]%
        {di2017gp}
\bibfield{author}{\bibinfo{person}{Xing Di}, \bibinfo{person}{Vishwanath~A
  Sindagi}, {and} \bibinfo{person}{Vishal~M Patel}.}
  \bibinfo{year}{2018}\natexlab{}.
\newblock \showarticletitle{GP-GAN: gender preserving GAN for synthesizing
  faces from landmarks}. In \bibinfo{booktitle}{{\em ICPR}}.
\newblock


\bibitem[\protect\citeauthoryear{Dolhansky and Ferrer}{Dolhansky and
  Ferrer}{2018}]%
        {dolhansky2017eye}
\bibfield{author}{\bibinfo{person}{Brian Dolhansky} {and}
  \bibinfo{person}{Cristian~Canton Ferrer}.} \bibinfo{year}{2018}\natexlab{}.
\newblock \showarticletitle{Eye In-Painting with Exemplar Generative
  Adversarial Networks}. In \bibinfo{booktitle}{{\em CVPR}}.
\newblock


\bibitem[\protect\citeauthoryear{Goodfellow, Pouget-Abadie, Mirza, Xu,
  Warde-Farley, Ozair, Courville, and Bengio}{Goodfellow et~al\mbox{.}}{2014}]%
        {goodfellow2014generative}
\bibfield{author}{\bibinfo{person}{Ian Goodfellow}, \bibinfo{person}{Jean
  Pouget-Abadie}, \bibinfo{person}{Mehdi Mirza}, \bibinfo{person}{Bing Xu},
  \bibinfo{person}{David Warde-Farley}, \bibinfo{person}{Sherjil Ozair},
  \bibinfo{person}{Aaron Courville}, {and} \bibinfo{person}{Yoshua Bengio}.}
  \bibinfo{year}{2014}\natexlab{}.
\newblock \showarticletitle{Generative adversarial nets}. In
  \bibinfo{booktitle}{{\em NIPS}}.
\newblock


\bibitem[\protect\citeauthoryear{He, Zhang, Ren, and Sun}{He
  et~al\mbox{.}}{2016}]%
        {he2016deep}
\bibfield{author}{\bibinfo{person}{Kaiming He}, \bibinfo{person}{Xiangyu
  Zhang}, \bibinfo{person}{Shaoqing Ren}, {and} \bibinfo{person}{Jian Sun}.}
  \bibinfo{year}{2016}\natexlab{}.
\newblock \showarticletitle{Deep residual learning for image recognition}. In
  \bibinfo{booktitle}{{\em CVPR}}.
\newblock


\bibitem[\protect\citeauthoryear{Heusel, Ramsauer, Unterthiner, Nessler,
  Klambauer, and Hochreiter}{Heusel et~al\mbox{.}}{2017}]%
        {heusel2017gans}
\bibfield{author}{\bibinfo{person}{Martin Heusel}, \bibinfo{person}{Hubert
  Ramsauer}, \bibinfo{person}{Thomas Unterthiner}, \bibinfo{person}{Bernhard
  Nessler}, \bibinfo{person}{G{\"u}nter Klambauer}, {and} \bibinfo{person}{Sepp
  Hochreiter}.} \bibinfo{year}{2017}\natexlab{}.
\newblock \showarticletitle{GANs trained by a two time-scale update rule
  converge to a Nash equilibrium}. In \bibinfo{booktitle}{{\em NIPS}}.
\newblock


\bibitem[\protect\citeauthoryear{Isola, Zhu, Zhou, and Efros}{Isola
  et~al\mbox{.}}{2017}]%
        {isola2017image}
\bibfield{author}{\bibinfo{person}{Phillip Isola}, \bibinfo{person}{Jun-Yan
  Zhu}, \bibinfo{person}{Tinghui Zhou}, {and} \bibinfo{person}{Alexei~A
  Efros}.} \bibinfo{year}{2017}\natexlab{}.
\newblock \showarticletitle{Image-to-image translation with conditional
  adversarial networks}. In \bibinfo{booktitle}{{\em CVPR}}.
\newblock


\bibitem[\protect\citeauthoryear{Johnson, Alahi, and Fei-Fei}{Johnson
  et~al\mbox{.}}{2016}]%
        {johnson2016perceptual}
\bibfield{author}{\bibinfo{person}{Justin Johnson}, \bibinfo{person}{Alexandre
  Alahi}, {and} \bibinfo{person}{Li Fei-Fei}.} \bibinfo{year}{2016}\natexlab{}.
\newblock \showarticletitle{Perceptual losses for real-time style transfer and
  super-resolution}. In \bibinfo{booktitle}{{\em ECCV}}.
\newblock


\bibitem[\protect\citeauthoryear{Karras, Aila, Laine, and Lehtinen}{Karras
  et~al\mbox{.}}{2018}]%
        {karras2017progressive}
\bibfield{author}{\bibinfo{person}{Tero Karras}, \bibinfo{person}{Timo Aila},
  \bibinfo{person}{Samuli Laine}, {and} \bibinfo{person}{Jaakko Lehtinen}.}
  \bibinfo{year}{2018}\natexlab{}.
\newblock \showarticletitle{Progressive growing of gans for improved quality,
  stability, and variation}. In \bibinfo{booktitle}{{\em ICLR}}.
\newblock


\bibitem[\protect\citeauthoryear{Kim, Cha, Kim, Lee, and Kim}{Kim
  et~al\mbox{.}}{2017}]%
        {kim2017learning}
\bibfield{author}{\bibinfo{person}{Taeksoo Kim}, \bibinfo{person}{Moonsu Cha},
  \bibinfo{person}{Hyunsoo Kim}, \bibinfo{person}{Jungkwon Lee}, {and}
  \bibinfo{person}{Jiwon Kim}.} \bibinfo{year}{2017}\natexlab{}.
\newblock \showarticletitle{Learning to discover cross-domain relations with
  generative adversarial networks}. In \bibinfo{booktitle}{{\em ICML}}.
\newblock


\bibitem[\protect\citeauthoryear{Kingma and Ba}{Kingma and Ba}{2015}]%
        {kingma2014adam}
\bibfield{author}{\bibinfo{person}{Diederik Kingma} {and}
  \bibinfo{person}{Jimmy Ba}.} \bibinfo{year}{2015}\natexlab{}.
\newblock \showarticletitle{Adam: A method for stochastic optimization}. In
  \bibinfo{booktitle}{{\em ICLR}}.
\newblock


\bibitem[\protect\citeauthoryear{Korshunova, Shi, Dambre, and Theis}{Korshunova
  et~al\mbox{.}}{2017}]%
        {korshunova2016fast}
\bibfield{author}{\bibinfo{person}{Iryna Korshunova}, \bibinfo{person}{Wenzhe
  Shi}, \bibinfo{person}{Joni Dambre}, {and} \bibinfo{person}{Lucas Theis}.}
  \bibinfo{year}{2017}\natexlab{}.
\newblock \showarticletitle{Fast face-swap using convolutional neural
  networks}. In \bibinfo{booktitle}{{\em ICCV}}.
\newblock


\bibitem[\protect\citeauthoryear{Ledig, Theis, Husz{\'a}r, Caballero,
  Cunningham, Acosta, Aitken, Tejani, Totz, Wang, et~al\mbox{.}}{Ledig
  et~al\mbox{.}}{2017}]%
        {ledig2016photo}
\bibfield{author}{\bibinfo{person}{Christian Ledig}, \bibinfo{person}{Lucas
  Theis}, \bibinfo{person}{Ferenc Husz{\'a}r}, \bibinfo{person}{Jose
  Caballero}, \bibinfo{person}{Andrew Cunningham}, \bibinfo{person}{Alejandro
  Acosta}, \bibinfo{person}{Andrew Aitken}, \bibinfo{person}{Alykhan Tejani},
  \bibinfo{person}{Johannes Totz}, \bibinfo{person}{Zehan Wang},
  {et~al\mbox{.}}} \bibinfo{year}{2017}\natexlab{}.
\newblock \showarticletitle{Photo-realistic single image super-resolution using
  a generative adversarial network}. In \bibinfo{booktitle}{{\em CVPR}}.
\newblock


\bibitem[\protect\citeauthoryear{Li and Wand}{Li and Wand}{2016}]%
        {li2016precomputed}
\bibfield{author}{\bibinfo{person}{Chuan Li} {and} \bibinfo{person}{Michael
  Wand}.} \bibinfo{year}{2016}\natexlab{}.
\newblock \showarticletitle{Precomputed real-time texture synthesis with
  markovian generative adversarial networks}. In \bibinfo{booktitle}{{\em
  ECCV}}.
\newblock


\bibitem[\protect\citeauthoryear{Li, Liang, Wei, Xu, Feng, and Yan}{Li
  et~al\mbox{.}}{2017a}]%
        {li2017perceptual}
\bibfield{author}{\bibinfo{person}{Jianan Li}, \bibinfo{person}{Xiaodan Liang},
  \bibinfo{person}{Yunchao Wei}, \bibinfo{person}{Tingfa Xu},
  \bibinfo{person}{Jiashi Feng}, {and} \bibinfo{person}{Shuicheng Yan}.}
  \bibinfo{year}{2017}\natexlab{a}.
\newblock \showarticletitle{Perceptual generative adversarial networks for
  small object detection}. In \bibinfo{booktitle}{{\em CVPR}}.
\newblock


\bibitem[\protect\citeauthoryear{Li, Liu, Yang, and Yang}{Li
  et~al\mbox{.}}{2017b}]%
        {li2017generative}
\bibfield{author}{\bibinfo{person}{Yijun Li}, \bibinfo{person}{Sifei Liu},
  \bibinfo{person}{Jimei Yang}, {and} \bibinfo{person}{Ming-Hsuan Yang}.}
  \bibinfo{year}{2017}\natexlab{b}.
\newblock \showarticletitle{Generative Face Completion}. In
  \bibinfo{booktitle}{{\em CVPR}}.
\newblock


\bibitem[\protect\citeauthoryear{Liu, Sun, Zhu, Bao, Wang, Shu, and Yan}{Liu
  et~al\mbox{.}}{2017}]%
        {liu2017face}
\bibfield{author}{\bibinfo{person}{Si Liu}, \bibinfo{person}{Yao Sun},
  \bibinfo{person}{Defa Zhu}, \bibinfo{person}{Renda Bao}, \bibinfo{person}{Wei
  Wang}, \bibinfo{person}{Xiangbo Shu}, {and} \bibinfo{person}{Shuicheng Yan}.}
  \bibinfo{year}{2017}\natexlab{}.
\newblock \showarticletitle{Face aging with contextual generative adversarial
  nets}. In \bibinfo{booktitle}{{\em ACM MM}}.
\newblock


\bibitem[\protect\citeauthoryear{Liu, Sun, Zhu, Ren, Chen, Feng, and Han}{Liu
  et~al\mbox{.}}{2018}]%
        {liu2018cross}
\bibfield{author}{\bibinfo{person}{Si Liu}, \bibinfo{person}{Yao Sun},
  \bibinfo{person}{Defa Zhu}, \bibinfo{person}{Guanghui Ren},
  \bibinfo{person}{Yu Chen}, \bibinfo{person}{Jiashi Feng}, {and}
  \bibinfo{person}{Jizhong Han}.} \bibinfo{year}{2018}\natexlab{}.
\newblock \showarticletitle{Cross-domain human parsing via adversarial feature
  and label adaptation}. In \bibinfo{booktitle}{{\em AAAI}}.
\newblock


\bibitem[\protect\citeauthoryear{Liu, Luo, Qiu, Wang, and Tang}{Liu
  et~al\mbox{.}}{2016}]%
        {liu2016deepfashion}
\bibfield{author}{\bibinfo{person}{Ziwei Liu}, \bibinfo{person}{Ping Luo},
  \bibinfo{person}{Shi Qiu}, \bibinfo{person}{Xiaogang Wang}, {and}
  \bibinfo{person}{Xiaoou Tang}.} \bibinfo{year}{2016}\natexlab{}.
\newblock \showarticletitle{Deepfashion: Powering robust clothes recognition
  and retrieval with rich annotations}. In \bibinfo{booktitle}{{\em CVPR}}.
\newblock


\bibitem[\protect\citeauthoryear{Luc, Couprie, Chintala, and Verbeek}{Luc
  et~al\mbox{.}}{2016}]%
        {luc2016semantic}
\bibfield{author}{\bibinfo{person}{Pauline Luc}, \bibinfo{person}{Camille
  Couprie}, \bibinfo{person}{Soumith Chintala}, {and} \bibinfo{person}{Jakob
  Verbeek}.} \bibinfo{year}{2016}\natexlab{}.
\newblock \showarticletitle{Semantic segmentation using adversarial networks}.
  In \bibinfo{booktitle}{{\em NIPS Workshops}}.
\newblock


\bibitem[\protect\citeauthoryear{Ma, Jia, Sun, Schiele, Tuytelaars, and
  Van~Gool}{Ma et~al\mbox{.}}{2017}]%
        {ma2017pose}
\bibfield{author}{\bibinfo{person}{Liqian Ma}, \bibinfo{person}{Xu Jia},
  \bibinfo{person}{Qianru Sun}, \bibinfo{person}{Bernt Schiele},
  \bibinfo{person}{Tinne Tuytelaars}, {and} \bibinfo{person}{Luc Van~Gool}.}
  \bibinfo{year}{2017}\natexlab{}.
\newblock \showarticletitle{Pose Guided Person Image Generation}. In
  \bibinfo{booktitle}{{\em NIPS}}.
\newblock


\bibitem[\protect\citeauthoryear{Ma, Sun, Georgoulis, Van~Gool, Schiele, and
  Fritz}{Ma et~al\mbox{.}}{2018}]%
        {ma2017disentangled}
\bibfield{author}{\bibinfo{person}{Liqian Ma}, \bibinfo{person}{Qianru Sun},
  \bibinfo{person}{Stamatios Georgoulis}, \bibinfo{person}{Luc Van~Gool},
  \bibinfo{person}{Bernt Schiele}, {and} \bibinfo{person}{Mario Fritz}.}
  \bibinfo{year}{2018}\natexlab{}.
\newblock \showarticletitle{Disentangled Person Image Generation}. In
  \bibinfo{booktitle}{{\em CVPR}}.
\newblock


\bibitem[\protect\citeauthoryear{Mathieu, Couprie, and LeCun}{Mathieu
  et~al\mbox{.}}{2016}]%
        {mathieu2015deep}
\bibfield{author}{\bibinfo{person}{Michael Mathieu}, \bibinfo{person}{Camille
  Couprie}, {and} \bibinfo{person}{Yann LeCun}.}
  \bibinfo{year}{2016}\natexlab{}.
\newblock \showarticletitle{Deep multi-scale video prediction beyond mean
  square error}.
\newblock \bibinfo{journal}{{\em ICLR\/}} (\bibinfo{year}{2016}).
\newblock


\bibitem[\protect\citeauthoryear{Memo and Zanuttigh}{Memo and
  Zanuttigh}{2016}]%
        {memo2016head}
\bibfield{author}{\bibinfo{person}{Alvise Memo} {and} \bibinfo{person}{Pietro
  Zanuttigh}.} \bibinfo{year}{2016}\natexlab{}.
\newblock \showarticletitle{Head-mounted gesture controlled interface for
  human-computer interaction}.
\newblock \bibinfo{journal}{{\em Springer Multimedia Tools and Applications\/}}
  (\bibinfo{year}{2016}), \bibinfo{pages}{1--27}.
\newblock


\bibitem[\protect\citeauthoryear{Mirza and Osindero}{Mirza and
  Osindero}{2014}]%
        {mirza2014conditional}
\bibfield{author}{\bibinfo{person}{Mehdi Mirza} {and} \bibinfo{person}{Simon
  Osindero}.} \bibinfo{year}{2014}\natexlab{}.
\newblock \showarticletitle{Conditional generative adversarial nets}.
\newblock \bibinfo{journal}{{\em arXiv preprint arXiv:1411.1784\/}}
  (\bibinfo{year}{2014}).
\newblock


\bibitem[\protect\citeauthoryear{Nguyen, Le, Vu, and Phung}{Nguyen
  et~al\mbox{.}}{2017}]%
        {nguyen2017dual}
\bibfield{author}{\bibinfo{person}{Tu Nguyen}, \bibinfo{person}{Trung Le},
  \bibinfo{person}{Hung Vu}, {and} \bibinfo{person}{Dinh Phung}.}
  \bibinfo{year}{2017}\natexlab{}.
\newblock \showarticletitle{Dual discriminator generative adversarial nets}. In
  \bibinfo{booktitle}{{\em NIPS}}.
\newblock


\bibitem[\protect\citeauthoryear{Oord, Dieleman, Zen, Simonyan, Vinyals,
  Graves, Kalchbrenner, Senior, and Kavukcuoglu}{Oord et~al\mbox{.}}{2016}]%
        {oord2016wavenet}
\bibfield{author}{\bibinfo{person}{Aaron van~den Oord}, \bibinfo{person}{Sander
  Dieleman}, \bibinfo{person}{Heiga Zen}, \bibinfo{person}{Karen Simonyan},
  \bibinfo{person}{Oriol Vinyals}, \bibinfo{person}{Alex Graves},
  \bibinfo{person}{Nal Kalchbrenner}, \bibinfo{person}{Andrew Senior}, {and}
  \bibinfo{person}{Koray Kavukcuoglu}.} \bibinfo{year}{2016}\natexlab{}.
\newblock \showarticletitle{Wavenet: A generative model for raw audio}. In
  \bibinfo{booktitle}{{\em SSW}}.
\newblock


\bibitem[\protect\citeauthoryear{Park, Yang, Yumer, Ceylan, and Berg}{Park
  et~al\mbox{.}}{2017}]%
        {park2017transformation}
\bibfield{author}{\bibinfo{person}{Eunbyung Park}, \bibinfo{person}{Jimei
  Yang}, \bibinfo{person}{Ersin Yumer}, \bibinfo{person}{Duygu Ceylan}, {and}
  \bibinfo{person}{Alexander~C Berg}.} \bibinfo{year}{2017}\natexlab{}.
\newblock \showarticletitle{Transformation-grounded image generation network
  for novel 3d view synthesis}. In \bibinfo{booktitle}{{\em CVPR}}.
\newblock


\bibitem[\protect\citeauthoryear{Pathak, Krahenbuhl, Donahue, Darrell, and
  Efros}{Pathak et~al\mbox{.}}{2016}]%
        {pathak2016context}
\bibfield{author}{\bibinfo{person}{Deepak Pathak}, \bibinfo{person}{Philipp
  Krahenbuhl}, \bibinfo{person}{Jeff Donahue}, \bibinfo{person}{Trevor
  Darrell}, {and} \bibinfo{person}{Alexei~A Efros}.}
  \bibinfo{year}{2016}\natexlab{}.
\newblock \showarticletitle{Context encoders: Feature learning by inpainting}.
  In \bibinfo{booktitle}{{\em CVPR}}.
\newblock


\bibitem[\protect\citeauthoryear{Perarnau, van~de Weijer, Raducanu, and
  {\'A}lvarez}{Perarnau et~al\mbox{.}}{2016}]%
        {perarnau2016invertible}
\bibfield{author}{\bibinfo{person}{Guim Perarnau}, \bibinfo{person}{Joost
  van~de Weijer}, \bibinfo{person}{Bogdan Raducanu}, {and}
  \bibinfo{person}{Jose~M {\'A}lvarez}.} \bibinfo{year}{2016}\natexlab{}.
\newblock \showarticletitle{Invertible Conditional GANs for image editing}. In
  \bibinfo{booktitle}{{\em NIPS Workshops}}.
\newblock


\bibitem[\protect\citeauthoryear{Qiao, Yao, Jiao, Li, Chen, and Wang}{Qiao
  et~al\mbox{.}}{2018}]%
        {qiao2018geometry}
\bibfield{author}{\bibinfo{person}{Fengchun Qiao}, \bibinfo{person}{Naiming
  Yao}, \bibinfo{person}{Zirui Jiao}, \bibinfo{person}{Zhihao Li},
  \bibinfo{person}{Hui Chen}, {and} \bibinfo{person}{Hongan Wang}.}
  \bibinfo{year}{2018}\natexlab{}.
\newblock \showarticletitle{Geometry-Contrastive Generative Adversarial Network
  for Facial Expression Synthesis}.
\newblock \bibinfo{journal}{{\em arXiv preprint arXiv:1802.01822\/}}
  (\bibinfo{year}{2018}).
\newblock


\bibitem[\protect\citeauthoryear{Reed, Akata, Yan, Logeswaran, Schiele, and
  Lee}{Reed et~al\mbox{.}}{2016b}]%
        {reed2016generative}
\bibfield{author}{\bibinfo{person}{Scott Reed}, \bibinfo{person}{Zeynep Akata},
  \bibinfo{person}{Xinchen Yan}, \bibinfo{person}{Lajanugen Logeswaran},
  \bibinfo{person}{Bernt Schiele}, {and} \bibinfo{person}{Honglak Lee}.}
  \bibinfo{year}{2016}\natexlab{b}.
\newblock \showarticletitle{Generative adversarial text to image synthesis}. In
  \bibinfo{booktitle}{{\em ICML}}.
\newblock


\bibitem[\protect\citeauthoryear{Reed, van~den Oord, Kalchbrenner, Bapst,
  Botvinick, and de~Freitas}{Reed et~al\mbox{.}}{2016c}]%
        {reed2016generating}
\bibfield{author}{\bibinfo{person}{Scott Reed}, \bibinfo{person}{A{\"a}ron
  van~den Oord}, \bibinfo{person}{Nal Kalchbrenner}, \bibinfo{person}{Victor
  Bapst}, \bibinfo{person}{Matt Botvinick}, {and} \bibinfo{person}{Nando de
  Freitas}.} \bibinfo{year}{2016}\natexlab{c}.
\newblock \showarticletitle{Generating interpretable images with controllable
  structure}.
\newblock \bibinfo{journal}{{\em Technical Report\/}} (\bibinfo{year}{2016}).
\newblock


\bibitem[\protect\citeauthoryear{Reed, Akata, Mohan, Tenka, Schiele, and
  Lee}{Reed et~al\mbox{.}}{2016a}]%
        {reed2016learning}
\bibfield{author}{\bibinfo{person}{Scott~E Reed}, \bibinfo{person}{Zeynep
  Akata}, \bibinfo{person}{Santosh Mohan}, \bibinfo{person}{Samuel Tenka},
  \bibinfo{person}{Bernt Schiele}, {and} \bibinfo{person}{Honglak Lee}.}
  \bibinfo{year}{2016}\natexlab{a}.
\newblock \showarticletitle{Learning what and where to draw}. In
  \bibinfo{booktitle}{{\em NIPS}}.
\newblock


\bibitem[\protect\citeauthoryear{Ren, Yuan, Meng, and Zhang}{Ren
  et~al\mbox{.}}{2013}]%
        {ren2013robust}
\bibfield{author}{\bibinfo{person}{Zhou Ren}, \bibinfo{person}{Junsong Yuan},
  \bibinfo{person}{Jingjing Meng}, {and} \bibinfo{person}{Zhengyou Zhang}.}
  \bibinfo{year}{2013}\natexlab{}.
\newblock \showarticletitle{Robust part-based hand gesture recognition using
  kinect sensor}.
\newblock \bibinfo{journal}{{\em IEEE TMM\/}} \bibinfo{volume}{15},
  \bibinfo{number}{5} (\bibinfo{year}{2013}), \bibinfo{pages}{1110--1120}.
\newblock


\bibitem[\protect\citeauthoryear{Salimans, Goodfellow, Zaremba, Cheung,
  Radford, and Chen}{Salimans et~al\mbox{.}}{2016}]%
        {salimans2016improved}
\bibfield{author}{\bibinfo{person}{Tim Salimans}, \bibinfo{person}{Ian
  Goodfellow}, \bibinfo{person}{Wojciech Zaremba}, \bibinfo{person}{Vicki
  Cheung}, \bibinfo{person}{Alec Radford}, {and} \bibinfo{person}{Xi Chen}.}
  \bibinfo{year}{2016}\natexlab{}.
\newblock \showarticletitle{Improved techniques for training gans}. In
  \bibinfo{booktitle}{{\em NIPS}}.
\newblock


\bibitem[\protect\citeauthoryear{Shu, Yumer, Hadap, Sunkavalli, Shechtman, and
  Samaras}{Shu et~al\mbox{.}}{2017}]%
        {shu2017neural}
\bibfield{author}{\bibinfo{person}{Zhixin Shu}, \bibinfo{person}{Ersin Yumer},
  \bibinfo{person}{Sunil Hadap}, \bibinfo{person}{Kalyan Sunkavalli},
  \bibinfo{person}{Eli Shechtman}, {and} \bibinfo{person}{Dimitris Samaras}.}
  \bibinfo{year}{2017}\natexlab{}.
\newblock \showarticletitle{Neural Face Editing with Intrinsic Image
  Disentangling}. In \bibinfo{booktitle}{{\em CVPR}}.
\newblock


\bibitem[\protect\citeauthoryear{Siarohin, Sangineto, Lathuiliere, and
  Sebe}{Siarohin et~al\mbox{.}}{2018}]%
        {siarohin2017deformable}
\bibfield{author}{\bibinfo{person}{Aliaksandr Siarohin}, \bibinfo{person}{Enver
  Sangineto}, \bibinfo{person}{Stephane Lathuiliere}, {and}
  \bibinfo{person}{Nicu Sebe}.} \bibinfo{year}{2018}\natexlab{}.
\newblock \showarticletitle{Deformable GANs for Pose-based Human Image
  Generation}. In \bibinfo{booktitle}{{\em CVPR}}.
\newblock


\bibitem[\protect\citeauthoryear{Simon, Joo, Matthews, and Sheikh}{Simon
  et~al\mbox{.}}{2017}]%
        {simon2017hand}
\bibfield{author}{\bibinfo{person}{Tomas Simon}, \bibinfo{person}{Hanbyul Joo},
  \bibinfo{person}{Iain Matthews}, {and} \bibinfo{person}{Yaser Sheikh}.}
  \bibinfo{year}{2017}\natexlab{}.
\newblock \showarticletitle{Hand Keypoint Detection in Single Images using
  Multiview Bootstrapping}. In \bibinfo{booktitle}{{\em CVPR}}.
\newblock


\bibitem[\protect\citeauthoryear{Simonyan and Zisserman}{Simonyan and
  Zisserman}{2015}]%
        {simonyan2014very}
\bibfield{author}{\bibinfo{person}{Karen Simonyan} {and}
  \bibinfo{person}{Andrew Zisserman}.} \bibinfo{year}{2015}\natexlab{}.
\newblock \showarticletitle{Very deep convolutional networks for large-scale
  image recognition}. In \bibinfo{booktitle}{{\em ICLR}}.
\newblock


\bibitem[\protect\citeauthoryear{Song, Lu, He, Sun, and Tan}{Song
  et~al\mbox{.}}{2017}]%
        {song2017geometry}
\bibfield{author}{\bibinfo{person}{Lingxiao Song}, \bibinfo{person}{Zhihe Lu},
  \bibinfo{person}{Ran He}, \bibinfo{person}{Zhenan Sun}, {and}
  \bibinfo{person}{Tieniu Tan}.} \bibinfo{year}{2017}\natexlab{}.
\newblock \showarticletitle{Geometry Guided Adversarial Facial Expression
  Synthesis}.
\newblock \bibinfo{journal}{{\em arXiv preprint arXiv:1712.03474\/}}
  (\bibinfo{year}{2017}).
\newblock


\bibitem[\protect\citeauthoryear{Sun, Ma, Oh, Van~Gool, Schiele, and Fritz}{Sun
  et~al\mbox{.}}{2018}]%
        {sun2017natural}
\bibfield{author}{\bibinfo{person}{Qianru Sun}, \bibinfo{person}{Liqian Ma},
  \bibinfo{person}{Seong~Joon Oh}, \bibinfo{person}{Luc Van~Gool},
  \bibinfo{person}{Bernt Schiele}, {and} \bibinfo{person}{Mario Fritz}.}
  \bibinfo{year}{2018}\natexlab{}.
\newblock \showarticletitle{Natural and Effective Obfuscation by Head
  Inpainting}. In \bibinfo{booktitle}{{\em CVPR}}.
\newblock


\bibitem[\protect\citeauthoryear{Taigman, Polyak, and Wolf}{Taigman
  et~al\mbox{.}}{2017}]%
        {taigman2016unsupervised}
\bibfield{author}{\bibinfo{person}{Yaniv Taigman}, \bibinfo{person}{Adam
  Polyak}, {and} \bibinfo{person}{Lior Wolf}.} \bibinfo{year}{2017}\natexlab{}.
\newblock \showarticletitle{Unsupervised cross-domain image generation}. In
  \bibinfo{booktitle}{{\em ICLR}}.
\newblock


\bibitem[\protect\citeauthoryear{Tang and Liu}{Tang and Liu}{2016}]%
        {tang2016novel}
\bibfield{author}{\bibinfo{person}{Hao Tang} {and} \bibinfo{person}{Hong Liu}.}
  \bibinfo{year}{2016}\natexlab{}.
\newblock \showarticletitle{A Novel Feature Matching Strategy for Large Scale
  Image Retrieval.}. In \bibinfo{booktitle}{{\em IJCAI}}.
\newblock


\bibitem[\protect\citeauthoryear{Tulyakov, Liu, Yang, and Kautz}{Tulyakov
  et~al\mbox{.}}{2018}]%
        {tulyakov2017mocogan}
\bibfield{author}{\bibinfo{person}{Sergey Tulyakov}, \bibinfo{person}{Ming-Yu
  Liu}, \bibinfo{person}{Xiaodong Yang}, {and} \bibinfo{person}{Jan Kautz}.}
  \bibinfo{year}{2018}\natexlab{}.
\newblock \showarticletitle{Mocogan: Decomposing motion and content for video
  generation}. In \bibinfo{booktitle}{{\em CVPR}}.
\newblock


\bibitem[\protect\citeauthoryear{Wang, Liu, Zhu, Tao, Kautz, and
  Catanzaro}{Wang et~al\mbox{.}}{2018}]%
        {wang2018high}
\bibfield{author}{\bibinfo{person}{Ting-Chun Wang}, \bibinfo{person}{Ming-Yu
  Liu}, \bibinfo{person}{Jun-Yan Zhu}, \bibinfo{person}{Andrew Tao},
  \bibinfo{person}{Jan Kautz}, {and} \bibinfo{person}{Bryan Catanzaro}.}
  \bibinfo{year}{2018}\natexlab{}.
\newblock \showarticletitle{High-resolution image synthesis and semantic
  manipulation with conditional gans}. In \bibinfo{booktitle}{{\em CVPR}}.
\newblock


\bibitem[\protect\citeauthoryear{Wang, Shrivastava, and Gupta}{Wang
  et~al\mbox{.}}{2017}]%
        {wang2017fast}
\bibfield{author}{\bibinfo{person}{Xiaolong Wang}, \bibinfo{person}{Abhinav
  Shrivastava}, {and} \bibinfo{person}{Abhinav Gupta}.}
  \bibinfo{year}{2017}\natexlab{}.
\newblock \showarticletitle{A-fast-rcnn: Hard positive generation via adversary
  for object detection}. In \bibinfo{booktitle}{{\em CVPR}}.
\newblock


\bibitem[\protect\citeauthoryear{Wei, Xavier, Dan, Elisa, Pascal, and Nicu}{Wei
  et~al\mbox{.}}{2018}]%
        {wei2017every}
\bibfield{author}{\bibinfo{person}{Wang Wei}, \bibinfo{person}{Alameda-Pineda
  Xavier}, \bibinfo{person}{Xu Dan}, \bibinfo{person}{Ricci Elisa},
  \bibinfo{person}{Fua Pascal}, {and} \bibinfo{person}{Sebe Nicu}.}
  \bibinfo{year}{2018}\natexlab{}.
\newblock \showarticletitle{Every Smile is Unique: Landmark-Guided Diverse
  Smile Generation}. In \bibinfo{booktitle}{{\em CVPR}}.
\newblock


\bibitem[\protect\citeauthoryear{Wu, Zheng, Zhang, and Huang}{Wu
  et~al\mbox{.}}{2017}]%
        {wu2017gp}
\bibfield{author}{\bibinfo{person}{Huikai Wu}, \bibinfo{person}{Shuai Zheng},
  \bibinfo{person}{Junge Zhang}, {and} \bibinfo{person}{Kaiqi Huang}.}
  \bibinfo{year}{2017}\natexlab{}.
\newblock \showarticletitle{Gp-gan: Towards realistic high-resolution image
  blending}.
\newblock \bibinfo{journal}{{\em arXiv preprint arXiv:1703.07195\/}}
  (\bibinfo{year}{2017}).
\newblock


\bibitem[\protect\citeauthoryear{Wu, Zhang, Xue, Freeman, and Tenenbaum}{Wu
  et~al\mbox{.}}{2016}]%
        {wu2016learning}
\bibfield{author}{\bibinfo{person}{Jiajun Wu}, \bibinfo{person}{Chengkai
  Zhang}, \bibinfo{person}{Tianfan Xue}, \bibinfo{person}{Bill Freeman}, {and}
  \bibinfo{person}{Josh Tenenbaum}.} \bibinfo{year}{2016}\natexlab{}.
\newblock \showarticletitle{Learning a probabilistic latent space of object
  shapes via 3d generative-adversarial modeling}. In \bibinfo{booktitle}{{\em
  NIPS}}.
\newblock


\bibitem[\protect\citeauthoryear{Xie, Dai, Du, Hovy, and Neubig}{Xie
  et~al\mbox{.}}{2017}]%
        {xie2017controllable}
\bibfield{author}{\bibinfo{person}{Qizhe Xie}, \bibinfo{person}{Zihang Dai},
  \bibinfo{person}{Yulun Du}, \bibinfo{person}{Eduard Hovy}, {and}
  \bibinfo{person}{Graham Neubig}.} \bibinfo{year}{2017}\natexlab{}.
\newblock \showarticletitle{Controllable Invariance through Adversarial Feature
  Learning}. In \bibinfo{booktitle}{{\em NIPS}}.
\newblock


\bibitem[\protect\citeauthoryear{Xu, Zhou, Zhang, and Yu}{Xu
  et~al\mbox{.}}{2017}]%
        {xu2017face}
\bibfield{author}{\bibinfo{person}{Runze Xu}, \bibinfo{person}{Zhiming Zhou},
  \bibinfo{person}{Weinan Zhang}, {and} \bibinfo{person}{Yong Yu}.}
  \bibinfo{year}{2017}\natexlab{}.
\newblock \showarticletitle{Face Transfer with Generative Adversarial Network}.
\newblock \bibinfo{journal}{{\em arXiv preprint arXiv:1710.06090\/}}
  (\bibinfo{year}{2017}).
\newblock


\bibitem[\protect\citeauthoryear{Yan, Xu, Ni, Zhang, and Yang}{Yan
  et~al\mbox{.}}{2017}]%
        {yan2017skeleton}
\bibfield{author}{\bibinfo{person}{Yichao Yan}, \bibinfo{person}{Jingwei Xu},
  \bibinfo{person}{Bingbing Ni}, \bibinfo{person}{Wendong Zhang}, {and}
  \bibinfo{person}{Xiaokang Yang}.} \bibinfo{year}{2017}\natexlab{}.
\newblock \showarticletitle{Skeleton-aided Articulated Motion Generation}. In
  \bibinfo{booktitle}{{\em ACM MM}}.
\newblock


\bibitem[\protect\citeauthoryear{Yang, Chou, and Yang}{Yang
  et~al\mbox{.}}{2017}]%
        {yang2017midinet}
\bibfield{author}{\bibinfo{person}{Li-Chia Yang}, \bibinfo{person}{Szu-Yu
  Chou}, {and} \bibinfo{person}{Yi-Hsuan Yang}.}
  \bibinfo{year}{2017}\natexlab{}.
\newblock \showarticletitle{MidiNet: A Convolutional Generative Adversarial
  Network for Symbolic-domain Music Generation}. In \bibinfo{booktitle}{{\em
  ISMIR}}.
\newblock


\bibitem[\protect\citeauthoryear{Yi, Zhang, Gong, et~al\mbox{.}}{Yi
  et~al\mbox{.}}{2017}]%
        {yi2017dualgan}
\bibfield{author}{\bibinfo{person}{Zili Yi}, \bibinfo{person}{Hao Zhang},
  \bibinfo{person}{Ping~Tan Gong}, {et~al\mbox{.}}}
  \bibinfo{year}{2017}\natexlab{}.
\newblock \showarticletitle{DualGAN: Unsupervised Dual Learning for
  Image-to-Image Translation}. In \bibinfo{booktitle}{{\em ICCV}}.
\newblock


\bibitem[\protect\citeauthoryear{Zhang, Xu, Li, Zhang, Wang, Huang, and
  Metaxas}{Zhang et~al\mbox{.}}{2017}]%
        {han2017stackgan}
\bibfield{author}{\bibinfo{person}{Han Zhang}, \bibinfo{person}{Tao Xu},
  \bibinfo{person}{Hongsheng Li}, \bibinfo{person}{Shaoting Zhang},
  \bibinfo{person}{Xiaogang Wang}, \bibinfo{person}{Xiaolei Huang}, {and}
  \bibinfo{person}{Dimitris Metaxas}.} \bibinfo{year}{2017}\natexlab{}.
\newblock \showarticletitle{StackGAN: Text to Photo-realistic Image Synthesis
  with Stacked Generative Adversarial Networks}. In \bibinfo{booktitle}{{\em
  {ICCV}}}.
\newblock


\bibitem[\protect\citeauthoryear{Zheng, Wang, Liu, and Tian}{Zheng
  et~al\mbox{.}}{2014a}]%
        {zheng2014packing}
\bibfield{author}{\bibinfo{person}{Liang Zheng}, \bibinfo{person}{Shengjin
  Wang}, \bibinfo{person}{Ziqiong Liu}, {and} \bibinfo{person}{Qi Tian}.}
  \bibinfo{year}{2014}\natexlab{a}.
\newblock \showarticletitle{Packing and padding: Coupled multi-index for
  accurate image retrieval}. In \bibinfo{booktitle}{{\em CVPR}}.
\newblock


\bibitem[\protect\citeauthoryear{Zheng, Wang, Zhou, and Tian}{Zheng
  et~al\mbox{.}}{2014b}]%
        {zheng2014bayes}
\bibfield{author}{\bibinfo{person}{Liang Zheng}, \bibinfo{person}{Shengjin
  Wang}, \bibinfo{person}{Wengang Zhou}, {and} \bibinfo{person}{Qi Tian}.}
  \bibinfo{year}{2014}\natexlab{b}.
\newblock \showarticletitle{Bayes merging of multiple vocabularies for scalable
  image retrieval}. In \bibinfo{booktitle}{{\em CVPR}}.
\newblock


\bibitem[\protect\citeauthoryear{Zhu, Park, Isola, and Efros}{Zhu
  et~al\mbox{.}}{2017a}]%
        {zhu2017unpaired}
\bibfield{author}{\bibinfo{person}{Jun-Yan Zhu}, \bibinfo{person}{Taesung
  Park}, \bibinfo{person}{Phillip Isola}, {and} \bibinfo{person}{Alexei~A
  Efros}.} \bibinfo{year}{2017}\natexlab{a}.
\newblock \showarticletitle{Unpaired image-to-image translation using
  cycle-consistent adversarial networks}. In \bibinfo{booktitle}{{\em ICCV}}.
\newblock


\bibitem[\protect\citeauthoryear{Zhu, Zhang, Pathak, Darrell, Efros, Wang, and
  Shechtman}{Zhu et~al\mbox{.}}{2017b}]%
        {zhu2017toward}
\bibfield{author}{\bibinfo{person}{Jun-Yan Zhu}, \bibinfo{person}{Richard
  Zhang}, \bibinfo{person}{Deepak Pathak}, \bibinfo{person}{Trevor Darrell},
  \bibinfo{person}{Alexei~A Efros}, \bibinfo{person}{Oliver Wang}, {and}
  \bibinfo{person}{Eli Shechtman}.} \bibinfo{year}{2017}\natexlab{b}.
\newblock \showarticletitle{Toward multimodal image-to-image translation}. In
  \bibinfo{booktitle}{{\em NIPS}}.
\newblock


\end{thebibliography}

\end{document}